\pdfoutput=1
\documentclass{article} 
\usepackage{iclr2017_conference,times}
\usepackage{url}

\usepackage[utf8]{inputenc}
\usepackage{hyperref}
\usepackage{amssymb}
\usepackage{amsmath}
\usepackage[normalem]{ulem}
\usepackage{dsfont}
\usepackage{enumitem}

\usepackage[algo2e, ruled]{algorithm2e}

\usepackage[center]{subfigure}

\usepackage{changepage}
\usepackage{graphicx, wrapfig}

\usepackage{pdfpages}

\usepackage{cancel}

\title{Stochastic Neural Networks for \\Hierarchical Reinforcement Learning}

\author{Carlos Florensa${^\dagger}$, Yan Duan${^\dagger}{^\ddagger}$, Pieter Abbeel${^\dagger}{^\ddagger}{^\mathsection}$ \\
$^\dagger$ UC Berkeley, Department of Electrical Engineering and Computer Science\\
$^\mathsection$ International Computer Science Institute\\
$^\ddagger$ OpenAI\\
\texttt{florensa@berkeley.edu, \{rocky,pieter\}@openai.com}
}

\newcommand{\sset}{\mathcal{S}}
\newcommand{\aset}{\mathcal{A}}

\newcommand{\mdpset}{\mathcal{M}}
\newcommand{\trans}{\mathcal{P}}
\newcommand{\agent}{\mathrm{agent}}
\iclrfinalcopy 

\begin{document}
	
	\maketitle

\begin{abstract}

Deep reinforcement learning has achieved many impressive results in recent years. However, tasks with sparse rewards or long horizons continue to pose significant challenges. To tackle these important problems, we propose a general framework that first learns useful skills in a pre-training environment, and then leverages the acquired skills for learning faster in downstream tasks.
Our approach brings together some of the strengths of intrinsic motivation and hierarchical methods: the learning of useful skill is guided by a single proxy reward, the design of which requires very minimal domain knowledge about the downstream tasks. Then a high-level policy is trained on top of these skills, providing a significant improvement of the exploration and allowing to tackle sparse rewards in the downstream tasks.
To efficiently pre-train a large span of skills, we use Stochastic Neural Networks combined with an information-theoretic regularizer. Our experiments\footnote{Code available at: \url{https://github.com/florensacc/snn4hrl}} show\footnote{\label{foot:video}Videos available at: \url{http://bit.ly/snn4hrl-videos}}  that this combination is effective in learning a wide span of interpretable skills in a sample-efficient way, and can significantly boost the learning performance uniformly across a wide range of downstream tasks.


\end{abstract}

\section{Introduction}

In recent years, deep reinforcement learning has achieved many impressive results, including playing Atari games from raw pixel inputs \citep{guo2014deep, mnih2015human, Schulman15TRPO}, mastering the game of Go \citep{silver2016mastering}, and acquiring advanced manipulation and locomotion skills from raw sensory inputs \citep{Schulman15TRPO, lillicrap2015continuous, watter2015embed, schulman2016high, heess2015learning, levine2016end}.
Despite these success stories, these deep RL algorithms typically employ naive exploration strategies such as $\epsilon$-greedy or uniform Gaussian exploration noise, which have been shown to perform poorly in tasks with sparse rewards \citep{duan2016benchmarking, houthooft2016variational, bellemare2016unifying}.
Tasks with sparse rewards are common in realistic scenarios. 
For example, in navigation tasks, the agent is not rewarded until it finds the target. The challenge is further complicated by long horizons, where naive exploration strategies can lead to exponentially large sample complexity \citep{osband2014generalization}.

To tackle these challenges, two main strategies have been pursued:
The first strategy is to design a hierarchy over the actions \citep{parr1998reinforcement, sutton1999between, dietterich2000hierarchical}.
By composing low-level actions into high-level primitives, the search space can be reduced exponentially. However, these approaches require domain-specific knowledge and careful hand-engineering. The second strategy uses intrinsic rewards to guide exploration \citep{schmidhuber1991curious, schmidhuber2010formal, houthooft2016variational, bellemare2016unifying}.
The computation of these intrinsic rewards does not require domain-specific knowledge.
However, when facing a collection of tasks, these methods do not provide a direct answer about how knowledge about solving one task may transfer to other tasks, and by solving each of the tasks from scratch, the overall sample complexity may still be high.

In this paper, we propose a general framework for training policies for a collection of tasks with sparse rewards.
Our framework first learns a span of skills in a pre-training environment, where it is employed with nothing but a proxy reward signal, whose design only requires very minimal domain knowledge about the downstream tasks. This proxy reward can be understood as a form of intrinsic motivation that encourages the agent to explore its own capabilities, without any goal information or sensor readings specific to each downstream task.
The set of skills can be used later in a wide collection of different tasks, by training a separate high-level policy for each task on top of the skills, thus reducing sample complexity uniformly.
To learn the span of skills, we propose to use Stochastic Neural Networks (SNNs) \citep{neal1990learning, neal1992connectionist, Tang2014_FSNN}, a class of neural networks with stochastic units in the computation graph.
This class of architectures can easily represent multi-modal policies, while achieving weight sharing among the different modes.
We parametrize the stochasticity in the network by feeding latent variables with simple distributions as extra input to the policy.
In the experiments, we observed that direct application of SNNs does not always guarantee that a wide range of skills will be learned. 
Hence, we propose an information-theoretic regularizer based on Mutual Information (MI) in the pre-training phase to encourage diversity of behaviors of the SNN policy.
Our experiments find that our hierarchical policy-learning framework can learn a wide range of skills, which are clearly interpretable as the latent code is varied. Furthermore, we show that training high-level policies on top of the learned skills can results in strong performance on a set of challenging tasks with long horizons and sparse rewards.

\section{Related Work}

One of the main appealing aspects of hierarchical reinforcement learning (HRL) is to use skills to reduce the search complexity of the problem \citep{parr1998reinforcement, sutton1999between, dietterich2000hierarchical}.
However, specifying a good hierarchy by hand requires domain-specific knowledge and careful engineering, hence motivating the need for learning skills automatically. Prior work on automatic learning of skills has largely focused on learning skills in discrete domains \citep{chentanez2004intrinsically, vigorito2010intrinsically}.
A popular approach there is to use statistics about state transitions to identify bottleneck states \citep{stolle2002learning, mannor2004dynamic, csimcsek2005identifying}.
It is however not clear how these techniques can be applied to settings with high-dimensional continuous spaces.
More recently, \cite{mnih2016strategic} propose a DRAW-like \citep{gregor2015draw} recurrent neural network architecture that can learn temporally extended macro actions.
However, the learning needs to be guided by external rewards as supervisory signals, and hence the algorithm cannot be straightforwardly applied in sparse reward settings.

There has also been work on learning skills in tasks with continuous actions \citep{schaal2005learning, konidaris2011autonomous, daniel2013autonomous, ranchod2015nonparametric}.
These methods extract useful skills from successful trajectories for the same (or closely related) tasks, and hence require first solving a comparably challenging task or demonstrations. 
Guided Policy Search (GPS)~\cite{levine2016end} leverages access to training in a simpler setting.  GPS trains with iLQG~\cite{todorov2005ilqg} in state space, and in parallel trains a neural net that only receives raw sensory inputs that has to agree with the iLQG controller.  The neural net policy is then able to generalize to new situations.

Another line of work on skill discovery particularly relevant to our approach is the HiREPS algorithm by \cite{daniel2012hierarchical} where, instead of having a mutual information bonus like in our proposed approach, they introduce a constraint on an equivalent metric. The solution approach is nevertheless very different as they cannot use policy gradients. Furthermore, although they do achieve multimodality like us, they only tried the episodic case where a single option is active during the rollout. Hence the hierarchical use of the learned skills is less clear. Similarly, the Option-critic architecture \citep{bacon2016option} can learn interpretable skills, but whether they can be reuse across complex tasks is still an open question. Recently, \cite{heess2016learning} have independently proposed to learn a range of skills in a pre-training environment that will be useful for the downstream tasks, which is similar to our framework.
However, their pre-training setup requires a set of goals to be specified.
In comparison, we use proxy rewards as the only signal to the agent during the pre-training phase, the construction of which only requires minimal domain knowledge or instrumentation of the environment.

\section{Preliminaries}
    
    We define a discrete-time finite-horizon discounted Markov decision process (MDP) by a tuple $M = (\sset, \aset, \trans, r, \rho_0, \gamma, T)$, in which $\sset$ is a state set, $\aset$ an action set, $\trans: \sset \times \aset \times \sset \rightarrow \mathbb{R}_{+}$ a transition probability distribution, $r: \sset \times \aset \rightarrow [-R_{\max}, R_{\max}]$ a bounded reward function, $\rho_0: \sset \to \mathbb{R}_+$ an initial state distribution, $\gamma \in [0, 1]$ a discount factor, and $T$ the horizon. When necessary, we attach a suffix to the symbols to resolve ambiguity, such as $\sset^M$. In policy search methods, we typically optimize a stochastic policy $\pi_{\theta}: \sset \times \aset \to \mathbb{R}_+$ parametrized by $\theta$. The objective is to maximize its expected discounted return, $ \eta(\pi_\theta) = \mathbb{E}_{\tau}[ \sum_{t=0}^T \gamma^t r(s_t, a_t) ]$, where $\tau = (s_0, a_0, \ldots)$ denotes the whole trajectory, $\displaystyle s_0 \sim \rho_0(s_0)$, $a_t \sim \pi_\theta(a_t|s_t)$, and $s_{t+1} \sim \trans(s_{t+1} | s_t, a_t)$.

\section{Problem statement}

Our main interest is in solving a collection of downstream tasks, specified via a collection of MDPs $\mdpset$. If these tasks do not share any common structure, we cannot expect to acquire a set of skills that will speed up learning for all of them. On the other hand, we want the structural assumptions to be minimal, to make our problem statement more generally applicable.

For each MDP $M \in \mdpset$, we assume that the state space $\sset^M$ can be factored into two components, $\sset_\agent$, and $\sset_{\mathrm{rest}}^M$, which only weakly interact with each other.
The $\sset_\agent$ should be the same for all MDPs in $\mdpset$. We also assume that all the MDPs share the same action space.
Intuitively, we consider a robot who faces a collection of tasks, where the dynamics of the robot are shared across tasks, and are covered by $\sset_\agent$, but there may be other components in a task-specific state space, which will be denoted by $\sset_{\mathrm{rest}}$. For instance, in a grasping task $M$, $\sset_{\mathrm{rest}}^M$ may include the positions of objects at interest or the new sensory input associated with the task. This specific structural assumption has been studied in the past as sharing the same \emph{agent-space} \citep{konidaris2007building}.

Given a collection of tasks satisfying the structural assumption, our objective for designing the algorithm is to minimize the total sample complexity required to solve these tasks. This has been more commonly studied in the past in sequential settings, where the agent is exposed to a sequence of tasks, and should learn to make use of experience gathered from solving earlier tasks to help solve later tasks \citep{taylor2009transfer, wilson2007multi, lazaric2010bayesian, devin2016learning}. However, this requires that the earlier tasks are relatively easy to solve (e.g., through good reward shaping or simply being easier) and is not directly applicable when all the tasks have sparse rewards, which is a much more challenging setting. In the next section, we describe our formulation, which takes advantage of a pre-training task that can be constructed with minimal domain knowledge, and can be applied to the more challenging scenario.

\section{Methodology}
\label{sec:method}

In this section, we describe our formulation to solve a collection of tasks, exploiting the structural assumption articulated above. In Sec.~\ref{section:method:pretraining}, we describe the pre-training environment, where we use proxy rewards to learn a useful span of skills. In Sec.~\ref{section:method:snn}, we motivate the usage of Stochastic Neural Networks (SNNs) and discuss the architectural design choices made to tailor them to skill learning for RL. 
In Sec.~\ref{section:method:inforeg}, we describe an information-theoretic regularizer that further improves the span of skills learned by SNNs. In Sec.~\ref{section:method:highlevel}, we describe the architecture of high-level policies over the learned skills, and the training procedure for the downstream tasks with sparse rewards. Finally, in Sec.~\ref{section:method:polopt}, we describe the policy optimization for both phases of training.

\subsection{Constructing the pre-training environment}
\label{section:method:pretraining}

Given a collection of tasks, we would like to construct a pre-training environment, where the agent can learn a span of skills, useful for enhancing exploration in downstream tasks. We achieve this by letting the agent freely interact with the environment in a minimal setup.
For a mobile robot, this can be a spacious environment where the robot can first learn the necessary locomotion skills; for a manipulator arm which will be used for object manipulation tasks, this can be an environment with many objects that the robot can interact with.

The skills learned in this environment will depend on the reward given to the agent.
Rather than setting different rewards in the pre-training environment corresponding to the desired skills, which requires precise specification about what each skill should entail, we use a generic proxy reward as the only reward signal to guide skill learning.
The design of the proxy reward should encourage the existence of locally optimal solutions, which will correspond to different skills the agent should learn. In other words, it encodes the prior knowledge about what high level behaviors might be useful in the downstream tasks, rewarding all of them roughly equally.
For a mobile robot, this reward can be as simple as proportional to the magnitude of the speed of the robot, without constraining the direction of movement.
For a manipulator arm, this reward can be the successful grasping of any object. 

Every time we train a usual uni-modal gaussian policy in such environment, it should converge towards a potentially different skill. As seen in Sec.\ \ref{section:results:skill_learning}, applying this approach to mobile robots gives a different direction (and type) of locomotion every time. But it is an inefficient procedure: training each policy from scratch is not sample efficient (sample complexity grows linearly with the number of skills to be learned) and nothing encourages the individual skills to be distinct. The first issue will be addressed in the next subsection by using Stochastic Neural Networks as policies and the second issue will be addressed in Sec.\ \ref{section:method:inforeg} by adding an information-theoretic regularizer.

\subsection{Stochastic Neural Networks for Skill Learning}
\label{section:method:snn}

To learn several skills at the same time, we propose to use Stochastic Neural Networks (SNNs), a general class of neural networks with stochastic units in the computation graph. There has been a large body of prior work on special classes of SNNs, such as Restricted Boltzmann Machines (RBMs) \citep{smolensky1986information, hinton2002training}, Deep Belief Networks (DBNs) \citep{hinton2006fast}, and Sigmoid Belief Networks (SBNs) \citep{neal1990learning, Tang2014_FSNN}. They have rich representation power and can in fact approximate any well-behaved probability distributions \citep{le2008representational, cho2013gaussian}. Policies modeled via SNNs can hence represent complex action distributions, especially multi-modal distributions.

For our purpose, we use a simple class of SNNs, where latent variables with fixed distributions are integrated with the inputs to the neural network (here, the observations from the environment) to form a joint embedding, which is then fed to a standard feed-forward neural network (FNN) with deterministic units, that computes distribution parameters for a uni-modal distribution (e.g. the mean and variance parameters of a multivariate Gaussian). We use simple categorical distributions with uniform weights for the latent variables, where the number of classes, $K$, is an hyperparameter that upper bounds the number of skills that we would like to learn.
\begin{figure}[th]
	\centering
		\vspace{-10pt}
	\subfigure[Concatenation]{
		\centering
		\label{fig:snn_architecture_concatenate}
		\includegraphics[width = 0.3\textwidth]{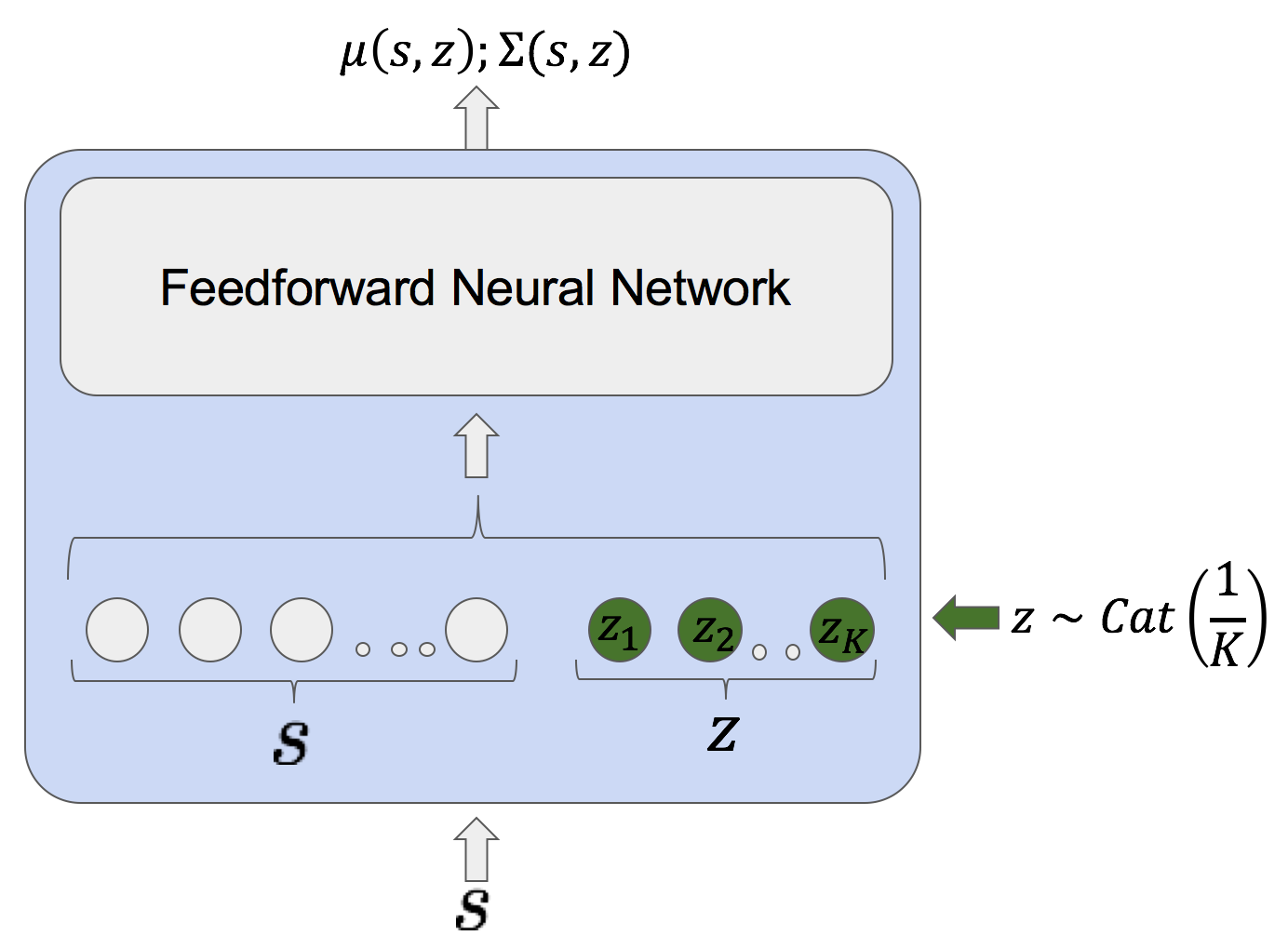}
	}
	\subfigure[Bilinear integration]{
		\centering
		\label{fig:snn_architecture_bilinear}
		\includegraphics[width = 0.5\textwidth]{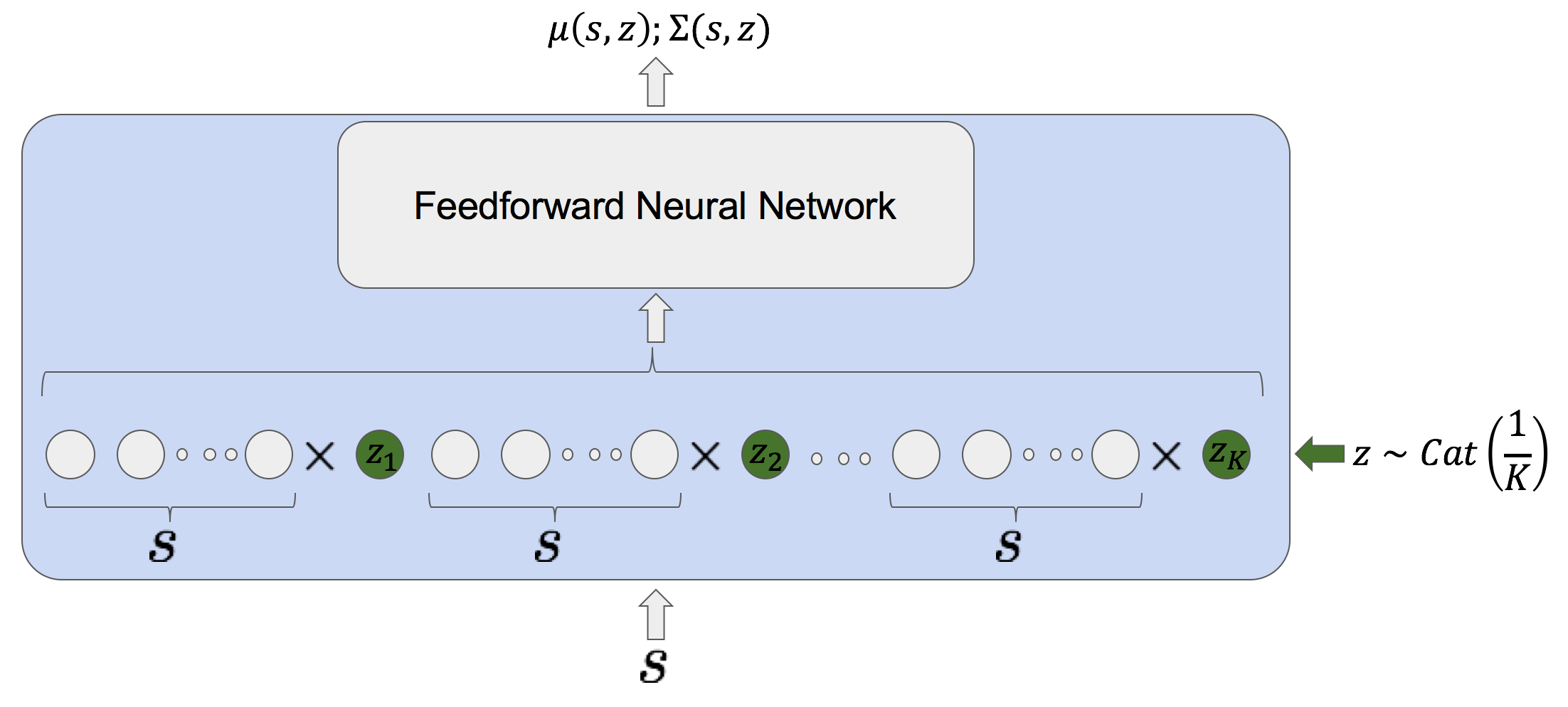}
	}
	\vspace{-10pt}
	\caption{Different architectures for the integration of the latent variables in a FNN}
	\label{fig:snn_architecture}
		\vspace{-10pt}
\end{figure}

The simplest joint embedding, as shown in Figure~\ref{fig:snn_architecture_concatenate}, is to concatenate the observations and the latent variables directly.\footnote{In scenarios where the observation is very high-dimensional such as images, we can form a low-dimensional embedding of the observation alone, say using a neural network, before jointly embedding it with the latent variable.} However, this limits the expressiveness power of integration between the observation and latent variable. Richer forms of integrations, such as multiplicative integrations and bilinear pooling, have been shown to have greater representation power and improve the optimization landscape, achieving better results when complex interactions are needed \citep{fukui2016multimodal, wu2016multiplicative}. Inspired by this work, we study using a simple bilinear integration, by forming the outer product between the observation and the latent variable (Fig.~\ref{fig:snn_architecture_bilinear}). Note that the concatenation integration effectively corresponds to changing the bias term of the first hidden layer depending on the latent code $z$ sampled, and the bilinear integration to changing all the first hidden layer weights. As shown in the experiments, the choice of integration greatly affects the quality of the span of skills that is learned. Our bilinear integration already yields a large span of skills, hence no other type of SNNs is studied in this work.

To obtain temporally extended and consistent behaviors associated with each latent code, in the pre-training environment we sample a latent code at the beginning of every rollout and keep it constant throughout the entire rollout.
After training, each of the latent codes in the categorical distribution will correspond to a different, interpretable skill, which can then be used for the downstream tasks.

Compared to training separate policies, training a single SNN allows for flexible weight-sharing schemes among different policies. It also takes a comparable amount of samples to train an SNN than a regular uni-modal gaussian policy, so the sample efficiency of training $K$ skills is effectively $K$ times better.
To further encourage the diversity of skills learned by the SNN we introduce an information theoretic regularizer, as detailed in the next section.

\subsection{Information-Theoretic Regularization}
\label{section:method:inforeg}

Although SNNs have sufficient expressiveness to represent multi-modal policies, there is nothing in the optimization that prevents them from collapsing into a single mode. We have not observed this worst case scenario in our experiments, but sometimes different latent codes correspond to very similar skills. It is desirable to have direct control over the diversity of skills that will be learned. To achieve this, we introduce an information-theoretic regularizer, inspired by recent success of similar objectives in encouraging interpretable representation learning in InfoGAN \cite{chen2016infogan}.

Concretely, we add an additional reward bonus, proportional to the mutual information (MI) between the latent variable and the current state. We only measure the MI with respect to a relevant subset of the state.
For a mobile robot, we choose this to be the $c=(x,y)$ coordinates of its center of mass (CoM). 
Formally, let $C$ be a random variable denoting the current CoM coordinate of the agent, and let $Z$ be the latent variable. Then the MI can be expressed as $I(Z;C) = H(Z) - H(Z|C)$, where $H$ denotes the entropy function. In our case, $H(Z)$ is constant since the probability distribution of the latent variable is fixed to uniform during this stage of training. Hence, maximizing the MI is equivalent to minimizing the conditional entropy $H(Z|C)$. As entropy is a measure of uncertainty, another interpretation of this bonus is that, given where the robot is, it should be easy to infer which skill the robot is currently performing. 
To penalize $H(Z|C)=-\mathbb{E}_{z,c}\log p(Z=z|C=c)$, we modify the reward received at every step as specified in Eq.\ \eqref{eq:MI_reward_modification}, where $\hat{p}(Z=z^n|c^n_t)$ is an estimate of the posterior probability of the latent code $z^n$ sampled on rollout $n$, given the coordinates $c^n_t$ at time $t$ of that rollout.
\begin{align}
\label{eq:MI_reward_modification}
    R^n_t\leftarrow R^n_t + \alpha_H~ \log\hat{p}(Z=z^n|c^n_t)
\end{align}
To estimate the posterior $\hat{p}(Z=z^n|c^n_t)$, we apply the following discretization: we partition the $(x,y)$ coordinate space into cells and map the continuous-valued CoM coordinates into the cell containing it. Overloading notation, now $c_t$ is a discrete variable indicating the cell where the CoM is at time $t$. This way, calculation of the empirical posterior only requires maintaining visitation counts $m_c(z)$ of how many times each cell $c$ is visited when latent code $z$ is sampled. Given that we use a batch policy optimization method, we use all trajectories of the current batch to compute all the $m_c(z)$ and estimate $\hat{p}(Z=z^n|c^n_t)$, as shown in Eq.\ \eqref{eq:posterior_count}. If $c$ is more high dimensional, the posterior $\hat{p}(Z=z^n|c^n_t)$ can also be estimated by fitting a MLP regressor by Maximum Likelihood.
\begin{align}
\label{eq:posterior_count}
    \hat{p}\big(Z=z\big|(x,y)\big)\approx\hat{p}(Z=z|c)=\frac{m_c(z)}{\sum_{z'}m_c(z')}
\end{align}

\subsection{Learning High-level Policies}
\label{section:method:highlevel}

Given a span of $K$ skills learned during the pre-training task, we now describe how to use them as basic building blocks for solving tasks where only sparse reward signals are provided. Instead of learning from scratch the low-level controls, we leverage the provided skills by freezing them and training a high-level policy (Manager Neural Network) that operates by selecting a skill and committing to it for a fixed amount of steps $\mathcal{T}$. The imposed temporal consistency and the quality of the skills (in our case given by optimizing the proxy reward in the pre-train environment) yield an enhanced exploration allowing to solve the downstream tasks with sparse rewards.

\begin{wrapfigure}{r}{5cm}
\includegraphics[width = 5cm]{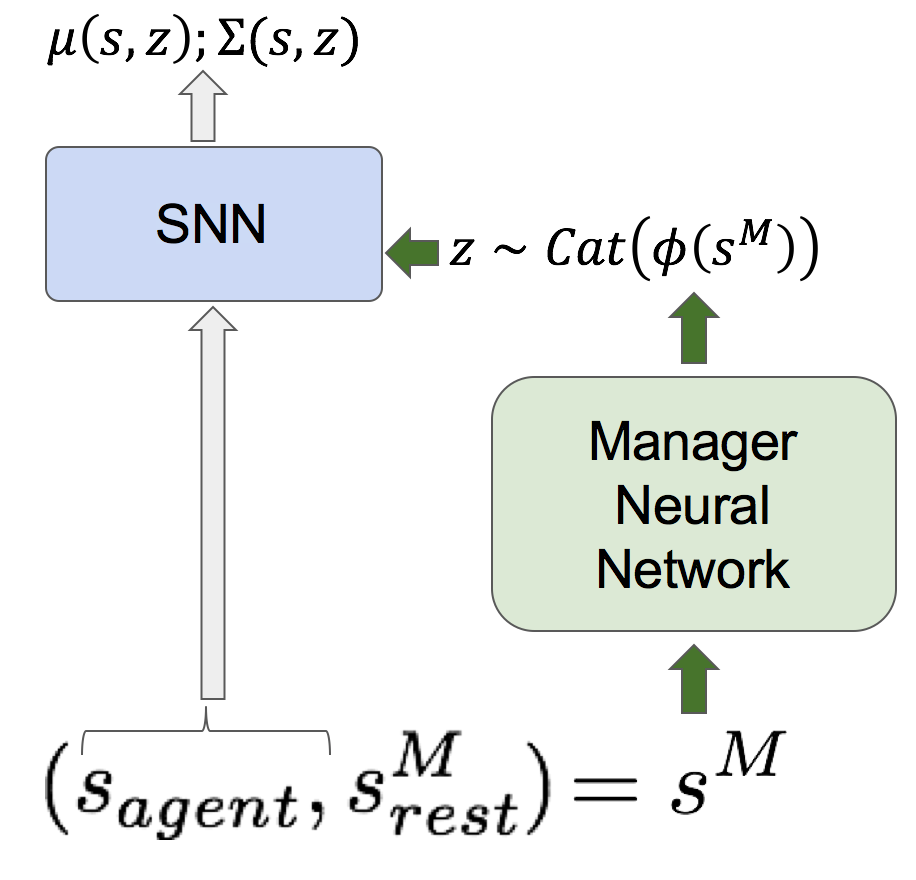}
\vspace{-20pt}
\caption{Hierarchical SNN architecture to solve downstream tasks}
\label{fig:hierarchical_snn_architecture}
\end{wrapfigure}

For any given task $M \in \mdpset$ we train a new Manager NN on top of the common skills. Given the factored representation of the state space $\sset^M$ as $\sset_\agent$ and $\sset_{\mathrm{rest}}^M$, the high-level policy receives the full state as input, and outputs the parametrization of a categorical distribution from which we sample a discrete action $z$ out of $K$ possible choices, corresponding to the $K$ available skills. If those skills are independently trained uni-modal policies, $z$ dictates the policy to use during the following $\mathcal{T}$ time-steps. If the skills are encapsulated in a SNN, $z$ is used in place of the latent variable. The architecture is depicted in Fig.\ \ref{fig:hierarchical_snn_architecture}. 

The weights of the low level and high level neural networks could also be jointly optimized to adapt the skills to the task at hand. This end-to-end training of a policy with discrete latent variables in the Stochastic Computation Graph could be done using straight-through estimators like the one proposed by \cite{jang2016gumbel} or \cite{maddison2016concrete}. Nevertheless, we show in our experiments that frozen low-level policies are already sufficient to achieve good performance in the studied downstream tasks, so these directions are left as future research.

\subsection{Policy Optimization}
\label{section:method:polopt}

For both the pre-training phase and the training of the high-level policies, we use Trust Region Policy Optimization (TRPO) as the policy optimization algorithm \citep{Schulman15TRPO}. We choose TRPO due to its excellent empirical performance and because it does not require excessive hyperparameter tuning. The training on downstream tasks does not require any modifications, except that the action space is now the set of skills that can be used. For the pre-training phase, due to the presence of categorical latent variables, the marginal distribution of $\pi(a|s)$ is now a mixture of Gaussians instead of a simple Gaussian, and it may even become intractable to compute if we use more complex latent variables. 
To avoid this issue, we consider the latent code as part of the observation. Given that $\pi(a|s,z)$ is still a Gaussian, TRPO can be applied without any modification.

\begin{algorithm2e}[H]
\SetAlgoLined
 \SetKwInOut{Input}{Input}
 \SetKwInOut{Output}{Output}
 \textbf{Initialize: }{Policy $\pi_{\theta}$; Latent dimension $K$}\;
 \While{Not trained}{
     \For{$n \leftarrow \ 1$ \KwTo $N$}{
        Sample $z_n\sim \text{Cat}\big(\frac{1}{K}\big)$\;
        Collect rollout with $z_n$ fixed\;
     }
     Compute $\hat{p}(Z=z|c)=\frac{m_c(z)}{\sum_{z'}m_c(z')}$\;
     Modify $R^n_t\leftarrow R^n_t + \alpha_H~ \log\hat{p}(Z=z^n|c^n_t)$\;
     Apply TRPO considering $z$ part of the observation\;
 }
 \caption{Skill training for SNNs with MI bonus}
 \label{alg:overall_MI}
\end{algorithm2e}

\section{Experiments}
\label{sec:experiments}
We have applied our framework to the two hierarchical tasks described in the benchmark by \cite{duan2016benchmarking}: Locomotion + Maze and Locomotion + Food Collection (Gather). The observation space of these tasks naturally decompose into $S_{agent}$ being the robot and $S_{rest}^M$ the task-specific attributes like walls, goals, and sensor readings. Here we report the results using the Swimmer robot, also described in the benchmark paper. In fact, the swimmer locomotion task described therein corresponds exactly to our pretrain task, as we also solely reward speed in a plain environment. We report results with more complex robots in Appendix \ref{sec:snake}-\ref{sec:ant}.

\begin{figure}[ht]
	\centering
	\subfigure[Maze 0]{
		\centering
		\label{fig:Maze0}
		\includegraphics[width = 0.2\textwidth, height=0.2\textwidth]{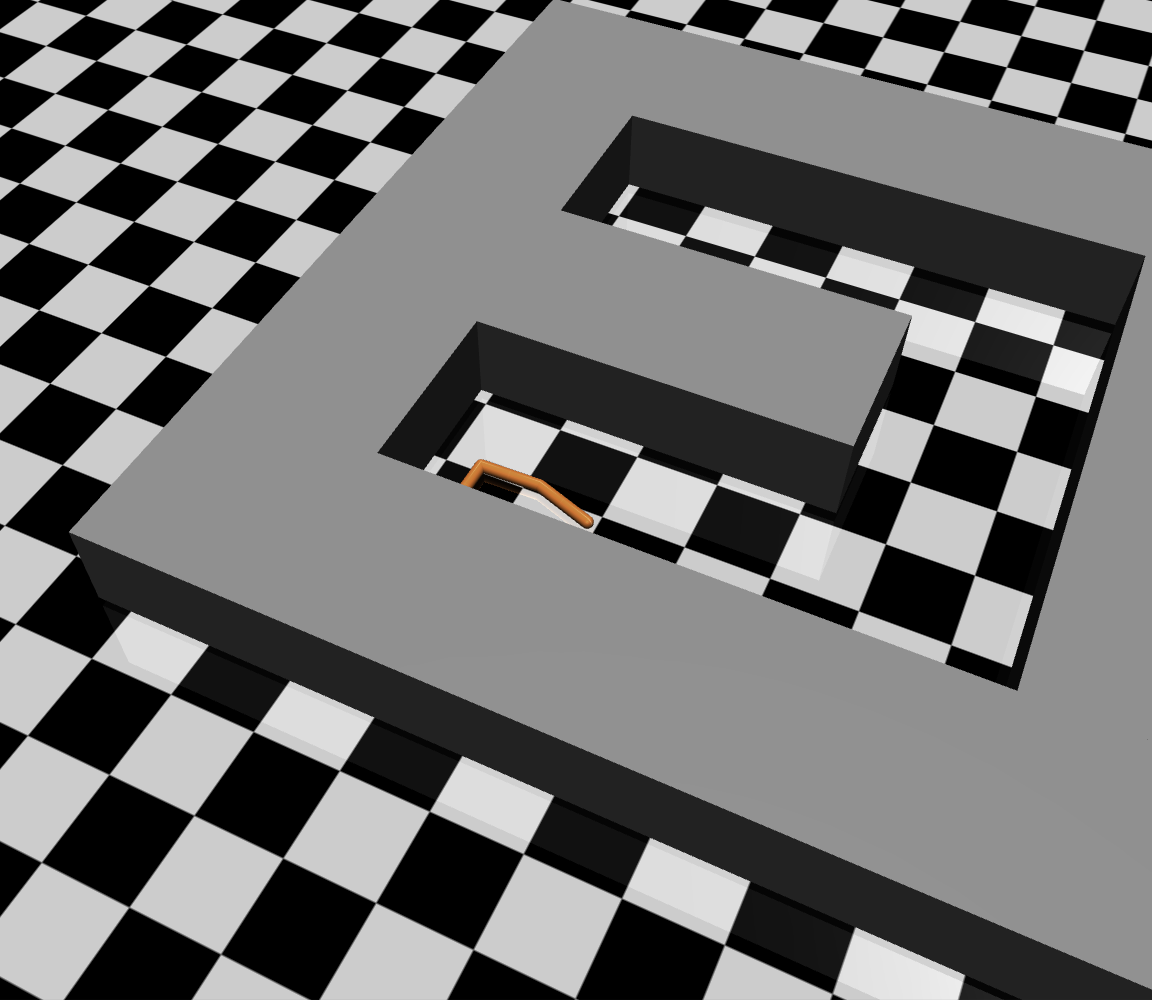}
	}
	\subfigure[Maze 1]{
		\centering
		\label{fig:Maze1}
		\includegraphics[width = 0.2\textwidth, height=0.2\textwidth]{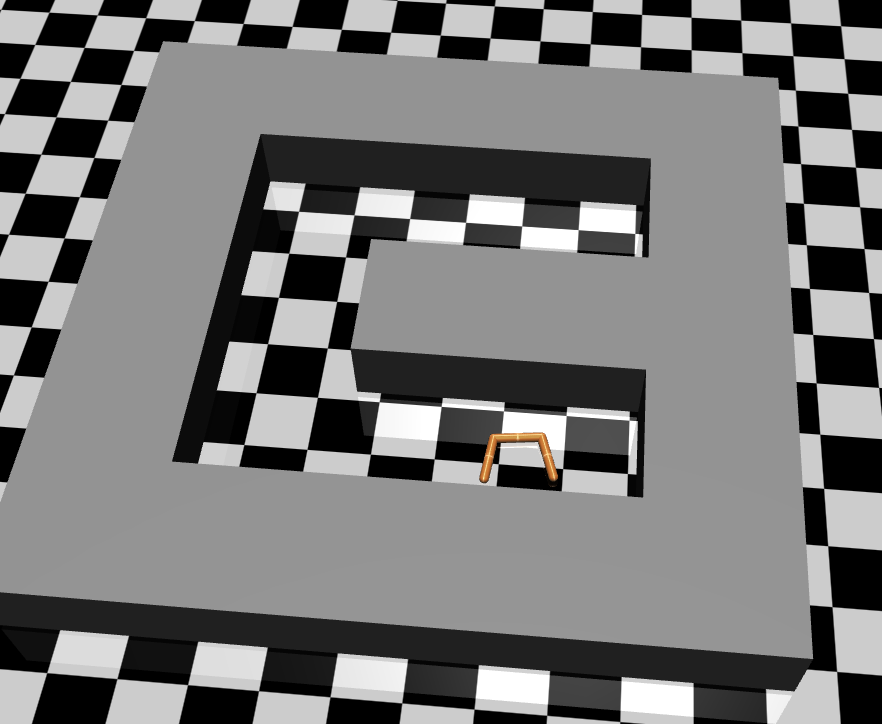}
	}
	\subfigure[Maze 2 or 3]{
		\centering
		\label{fig:Maze2}
		\includegraphics[width = 0.2\textwidth, height=0.2\textwidth]{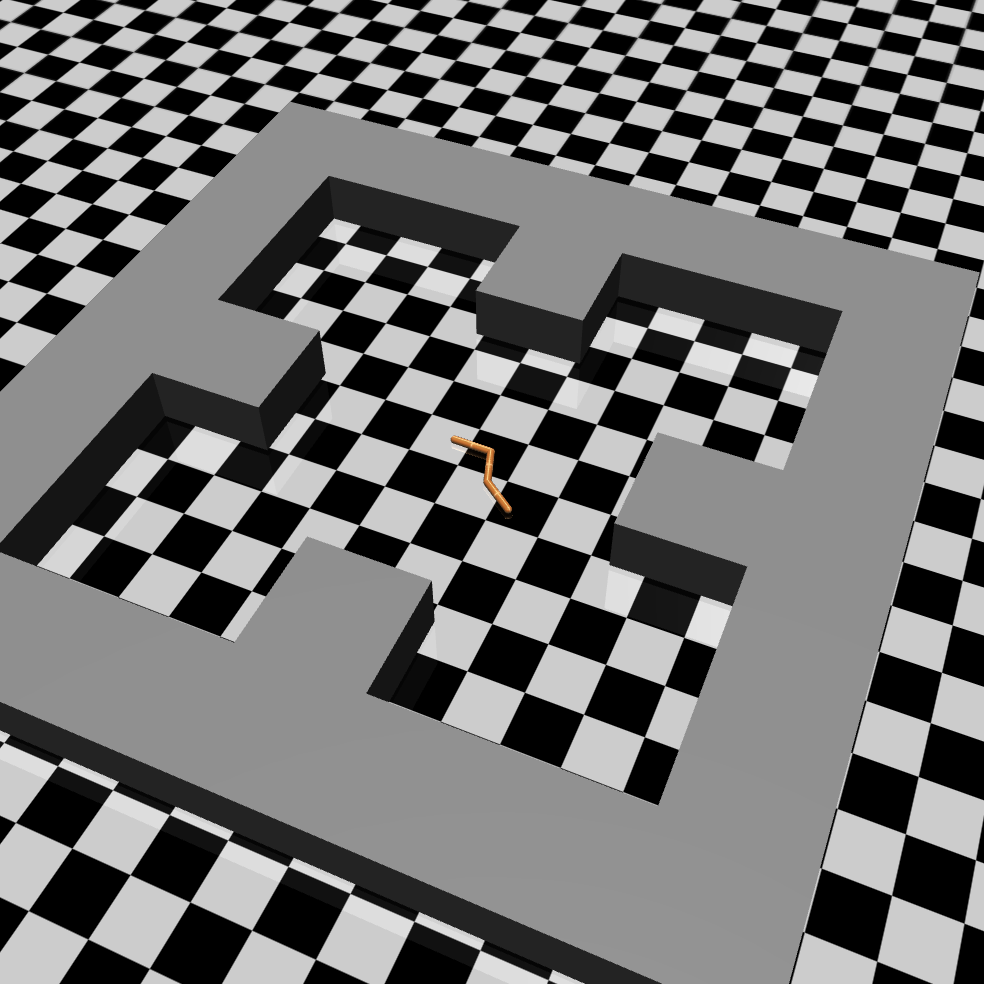}
	}
	\subfigure[Food Gather]{
		\centering
		\label{fig:FoodGather}
		\includegraphics[width = 0.2\textwidth, height=0.2\textwidth]{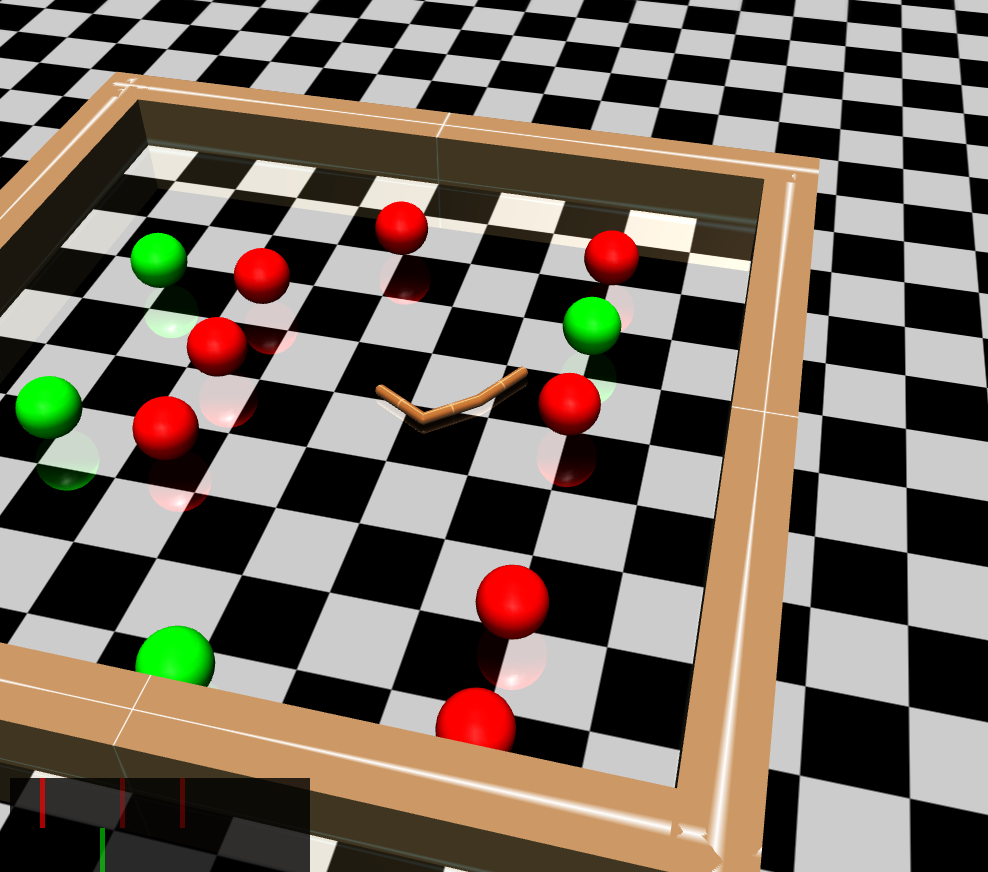}
	}
	\vspace{-10pt}
	\caption{Illustration of the sparse reward tasks}
	\label{fig:snn_multimodal_MI}
\end{figure}

To increase the variety of downstream tasks, we have constructed four different mazes. Maze 0 is the same as the one described in the benchmark \citep{duan2016benchmarking} and Maze 1 is its reflection, where the robot has to go backwards-right-right instead of forward-left-left. These correspond to  Figs.\ \ref{fig:Maze0}-\ref{fig:Maze1}, where the robot is shown in its starting position and has to reach the other end of the U turn. Mazes 2 and 3 are different instantiations of the environment shown in Fig.\ \ref{fig:Maze2}, where the goal has been placed in the North-East or in the South-West corner respectively. A reward of 1 is only granted when the robot reaches the goal position. In the Gather task depicted in Fig.\ \ref{fig:FoodGather}, the robot gets a reward of 1 for collecting green balls and a reward of -1 for the red ones, all of which are positioned randomly at the beginning of each episode. Apart from the robot joints positions and velocities, in these tasks the agent also receives LIDAR-like sensor readings of the distance to walls, goals or balls that are within a certain range. 

In the benchmark of continuous control problems \citep{duan2016benchmarking} it was shown that algorithms that employ naive exploration strategies could not solve them. More advanced intrinsically motivated explorations \citep{houthooft2016variational} do achieve some progress, and we report our stronger results with the exact same setting in Appendix \ref{sec:gather_compare}. Our hyperparameters for the neural network architectures and algorithms are detailed in the Appendix \ref{sec:hyper} and the full code is available\footnote{Code available at: \url{https://github.com/florensacc/snn4hrl}}.

\section{Results}
\label{sec:results}
We evaluate every step of the skill learning process, showing the relevance of the different pieces of our architecture and how they impact the exploration achieved when using them in a hierarchical fashion. Then we report the results\footnote{Videos available at: \url{http://bit.ly/snn4hrl-videos}} on the sparse environments described above. We seek to answer the following questions:
\begin{itemize}
\vspace{-8pt}
    \setlength\itemsep{0em}
    \item Can the multimodality of SNNs and the MI bonus consistently yield a large span of skills? 
    \item Can the pre-training experience improve the exploration in downstream environments?
    \item Does the enhanced exploration help to efficiently solve sparse complex tasks?
\end{itemize} 

\subsection{Skill learning in pretrain}
\label{section:results:skill_learning}

\begin{figure}[h!]
    \vspace{-10pt}
	\centering
	\subfigure[Independently trained policies in the pre-train MDP with the proxy reward of the CoM speed norm]{
		\centering
		\label{fig:visit-trpo-individuals}
		\includegraphics[trim={2cm 0 2cm 1.5cm}, clip, width=0.1\textwidth]{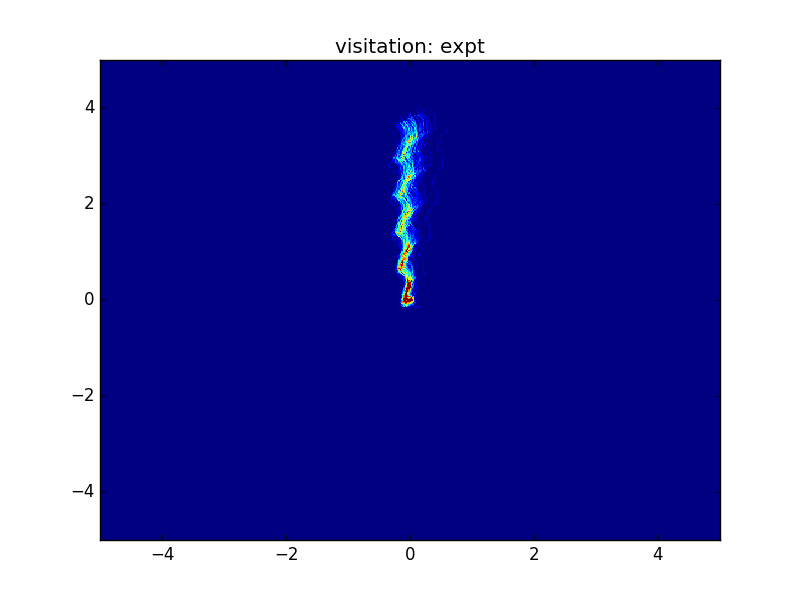}
	    \includegraphics[trim={2cm 0 2cm 1.5cm}, clip, width=0.1\textwidth]{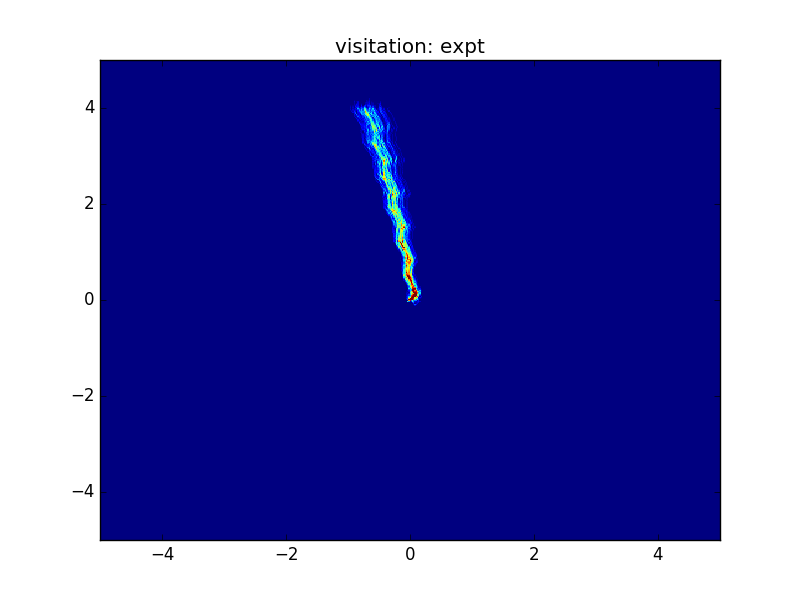}
		\includegraphics[trim={2cm 0 2cm 1.5cm}, clip, width=0.1\textwidth]{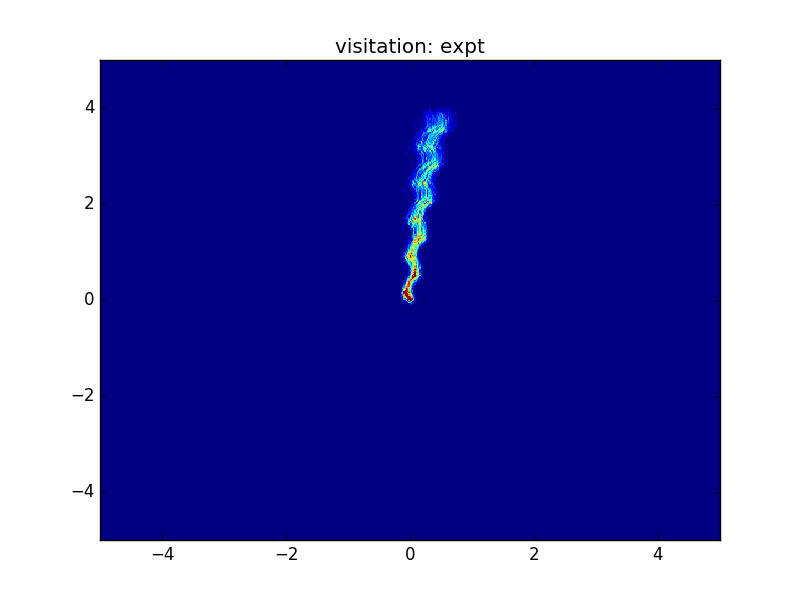}
	    \includegraphics[trim={2cm 0 2cm 1.5cm}, clip, width=0.1\textwidth]{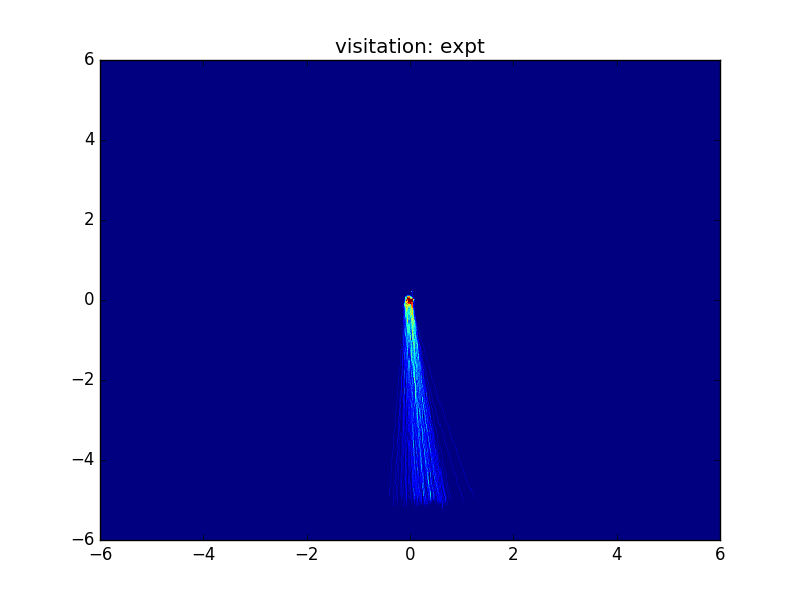}
		\includegraphics[trim={2cm 0 2cm 1.5cm}, clip, width=0.1\textwidth]{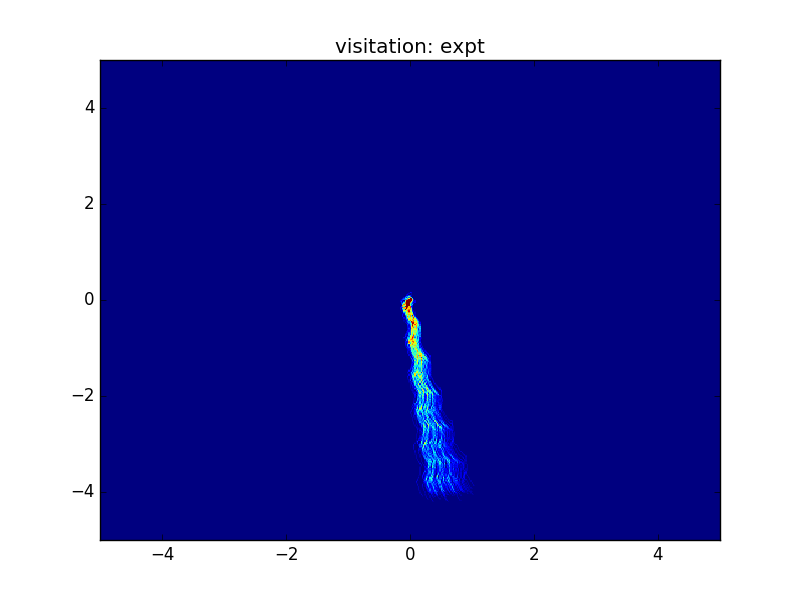}
	    \includegraphics[trim={2cm 0 2cm 1.5cm}, clip, width=0.1\textwidth]{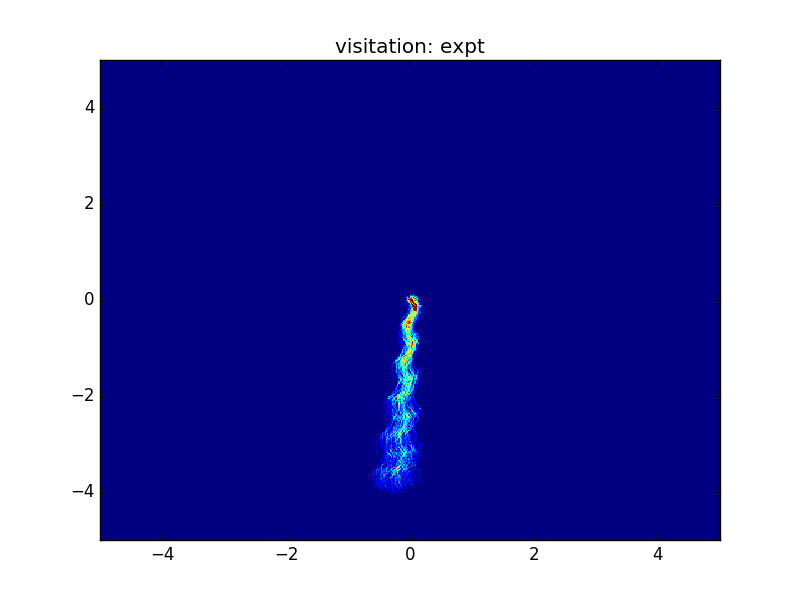}
	}
	\subfigure[Superposed policy visitations from (a)]{
		\centering
		\label{fig:visit_trpo6_likeSNN}
	    \includegraphics[trim={2cm 0 3.6cm 1.5cm}, clip, width=0.2\textwidth]{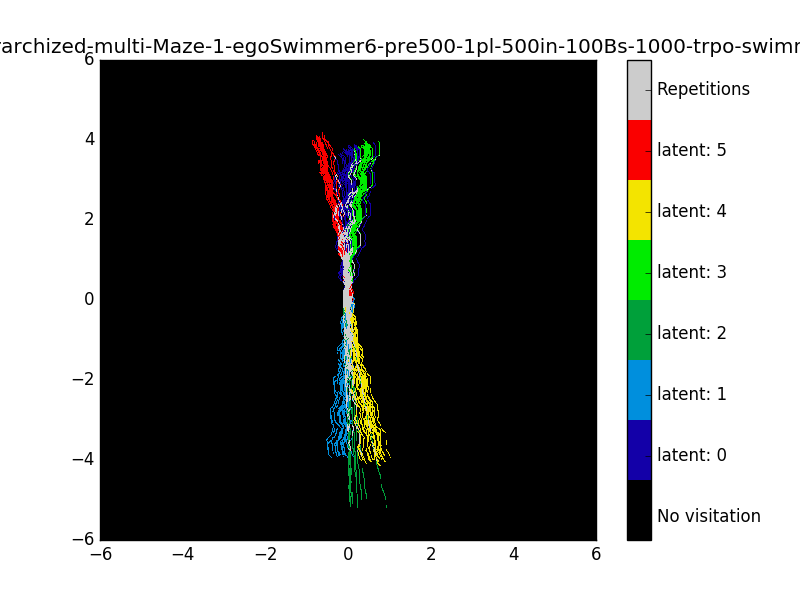}
	}
	\subfigure[SNN \emph{without} bilinear integration and increasing $\alpha_H=0,0.001,0.01,0.1$]{
		\centering
		\label{fig:visit_snn6_noBilinear}
		\includegraphics[trim={2cm 0 1cm 1.5cm}, clip, width = 0.2\textwidth]{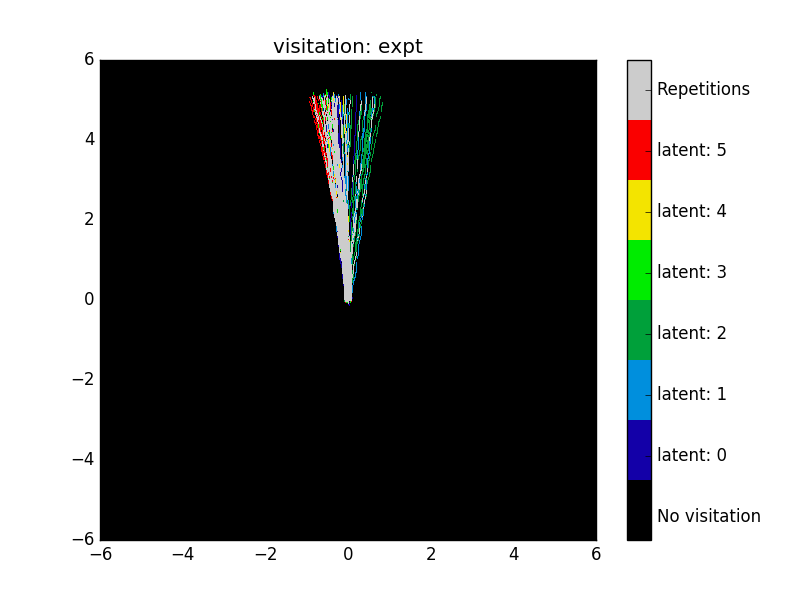}
		\includegraphics[trim={2cm 0 1cm 1.5cm}, clip, width = 0.2\textwidth]{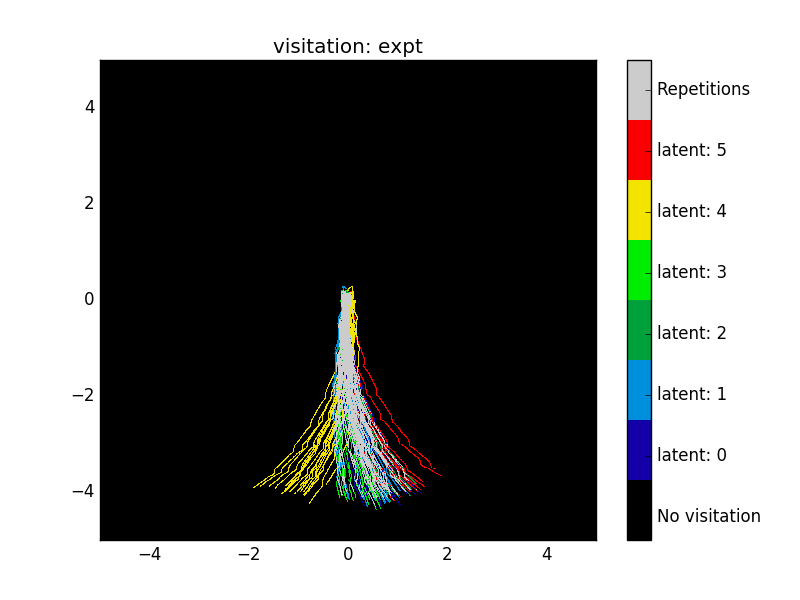}
		\includegraphics[trim={2cm 0 1cm 1.5cm}, clip, width = 0.2\textwidth]{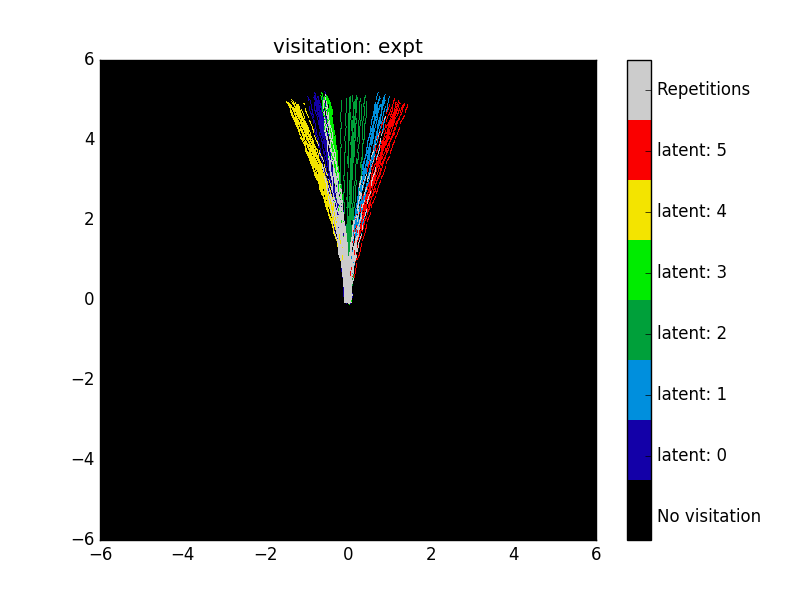}
		\includegraphics[trim={2cm 0 1cm 1.5cm}, clip, width = 0.2\textwidth]{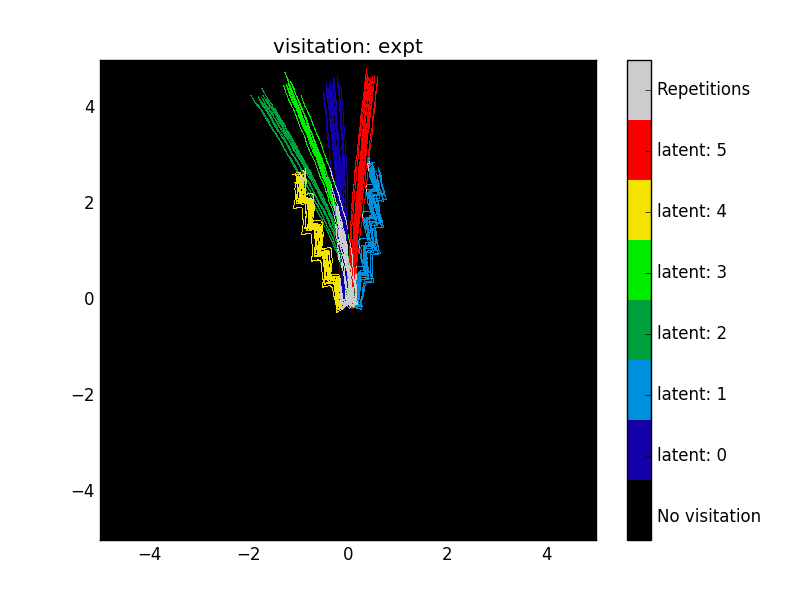}
	}
	\subfigure[SNN \emph{with} bilinear integration and increasing $\alpha_H=0,0.001,0.01,0.1$]{
		\centering
		\label{fig:visit_snn6_bilinear}
		\includegraphics[trim={2cm 0 1cm 1.5cm}, clip, width = 0.2\textwidth]{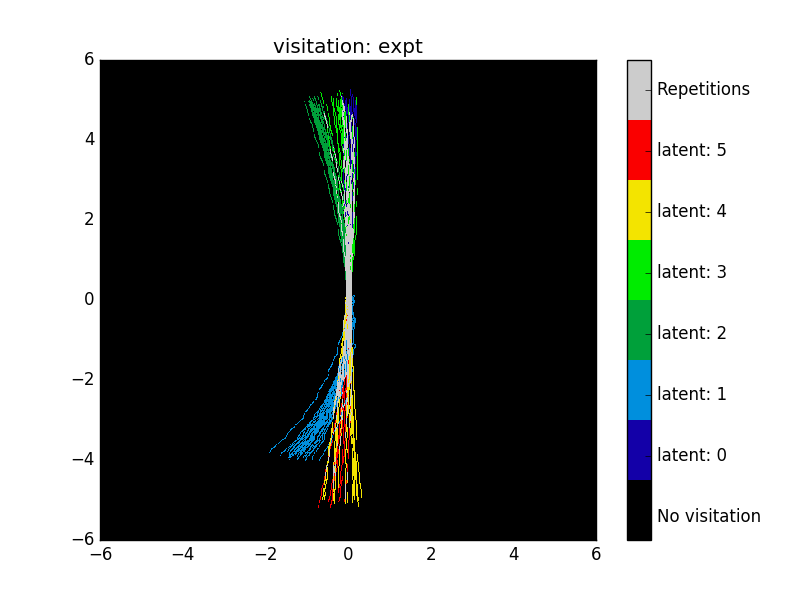}
		\includegraphics[trim={2cm 0 1cm 1.5cm}, clip, width = 0.2\textwidth]{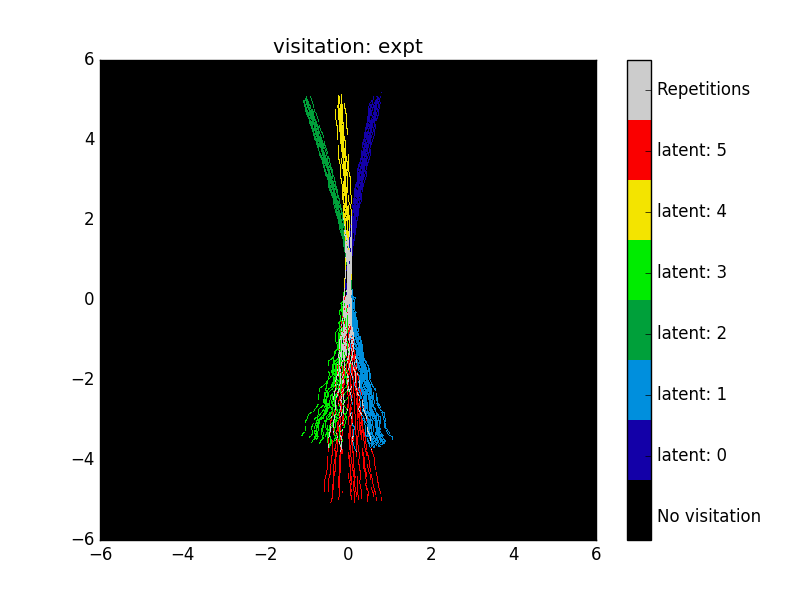}
		\includegraphics[trim={2cm 0 1cm 1.5cm}, clip, width = 0.2\textwidth]{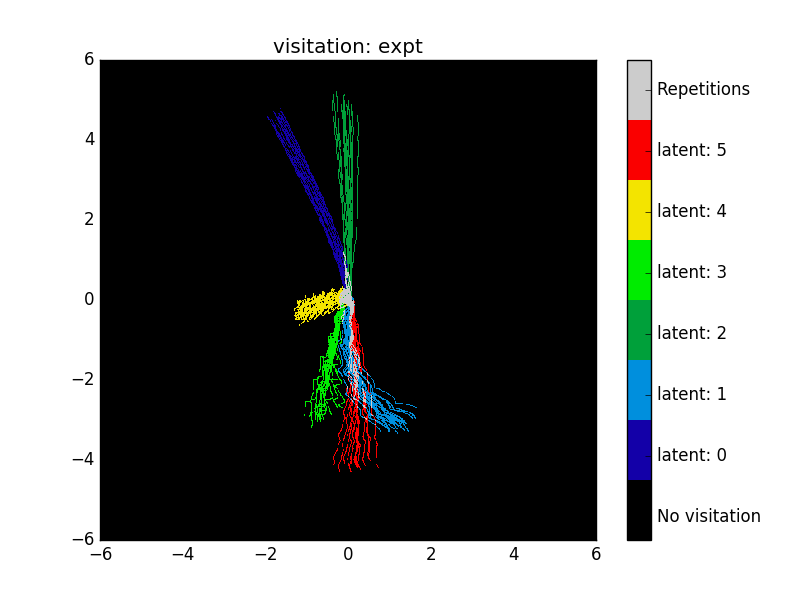}
		\includegraphics[trim={2cm 0 1cm 1.5cm}, clip, width = 0.2\textwidth]{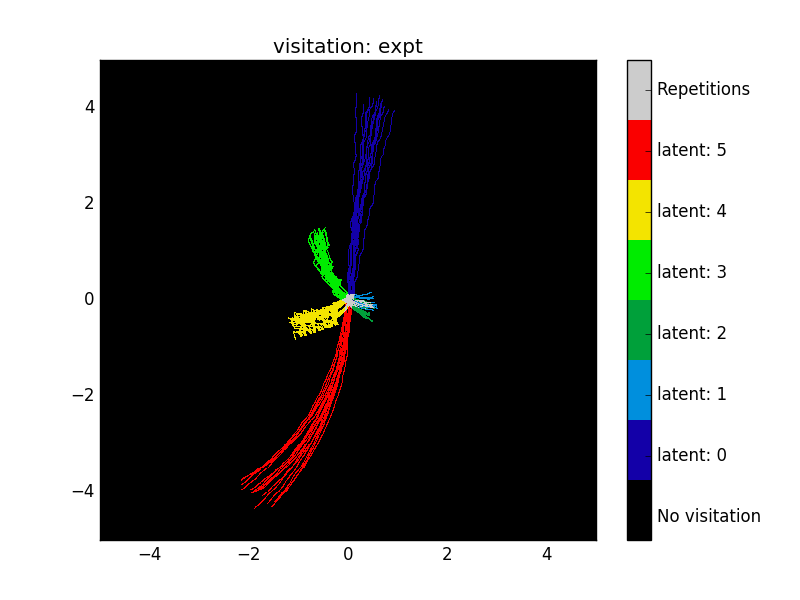}
	}
	\vspace{-10pt}
	\caption{Span of skills learn by different methods and architectures}
	\label{fig:visit_methods}
\end{figure}

To evaluate the diversity of the learned skills we use ``visitation plots'', showing the $(x,y)$ position of the robot's Center of Mass (CoM) during 100 rollouts of 500 time-steps each. At the beginning of every rollout, we reset the robot to the origin, always in the same orientation (as done during training). In Fig.\ \ref{fig:visit-trpo-individuals} we show the visitation plot of six different feed-forward policies, each trained from scratch in our pre-training environment. For better graphical interpretation and comparison with the next plots of the SNN policies, Fig.\ \ref{fig:visit_trpo6_likeSNN} superposes a batch of 50 rollout for each of the 6 policies, each with a different color. Given the morphology of the swimmer, it has a natural preference for forward and backward motion. Therefore, when no extra incentive is added, the visitation concentrates heavily on the direction it is always initialized with. Note nevertheless that the proxy reward is general enough so that each independently trained policy yields a different way
of advancing, hence granting a potentially useful skills to solve downstream tasks when embedded in our described Multi-policy hierarchical architecture.

Next we show that, using SNNs, we can learn a similar or even larger span of skills without training several independent policies. The number of samples used to train a SNN is the same as a single feed-forward policy from Fig.\ \ref{fig:visit-trpo-individuals}, therefore this method of learning skills effectively reduces the sample complexity by a factor equal to the numbers of skills being learned. With an adequate architecture and MI bonus, we show that the span of skills generated is also richer.   
In the two rows of Figs.\ \ref{fig:visit_snn6_noBilinear}-\ref{fig:visit_snn6_bilinear} we present the visitation plots of SNNs policies obtained for different design choices. Now the colors indicate the latent code that was sampled at the beginning of the rollout. We observe that each latent code generates a particular, interpretable behavior. Given that the initialization orientation is always the same, the different skills are truly distinct ways of moving: forward, backwards, or sideways. In the following we analyze the impact on the span of skills of the integration of latent variables (concatenation in first row and bilinear in the second) and the MI bonus (increasing coeficient $\alpha_H$ towards the right). Simple concatenation of latent variables with the observations rarely yields distinctive behaviors for each latent code sampled. On the other hand, we have observed that 80\% of the trained SNNs with bilinear integration acquire at least forward and backward motion associated with different latent codes. This can be further improved to 100\% by increasing $\alpha_H$ and we also observe that the MI bonus yields less overlapped skills and new advancing/turning skills, independently of the integration of the latent variables.

\subsection{Hierarchical use of skills}
\label{section:results:hierarchical_use_of_skills}

The hierarchical architectures we propose have a direct impact on the areas covered by random exploration. We will illustrate it with plots showing the visitation of the $(x-y)$ position of the Center of Mass (CoM) during single rollouts of one million steps, each generated with a different architecture. 

\begin{figure}[h!]
	\centering
	\subfigure[Gaussian noise with covariance $\Sigma=I$]{
		\centering
		\label{fig:visit-trpo-1M}
		\includegraphics[trim={0cm 0 0cm 1.2cm}, clip, width = 0.22\textwidth]{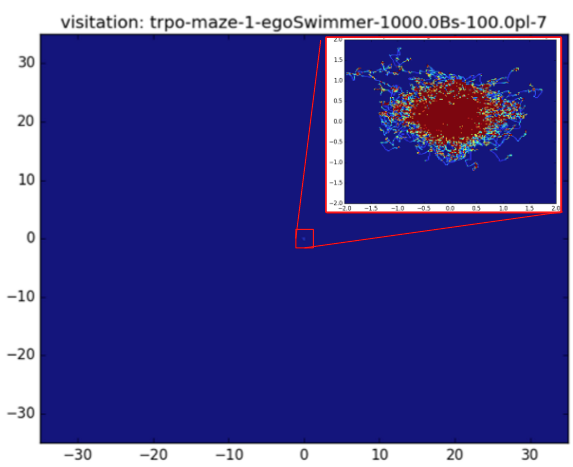}
	}
	\subfigure[Hierarchy with \textit{Multi-policy}]{
		\centering
		\label{fig:visit-hier-multi-1M}
		\includegraphics[trim={2cm 0.7cm 1cm 1.5cm}, clip, width = 0.22\textwidth]{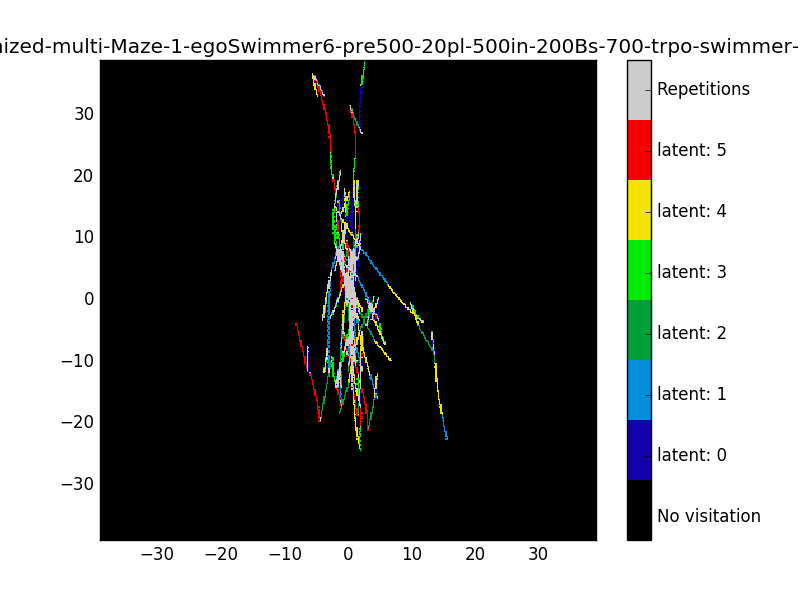}
	}
	\subfigure[Hierarchy with Bil-SNN $\alpha_H=0$ ]{
		\centering
		\label{fig:visit-hier-snnBil-1M}
		\includegraphics[trim={2cm 0.7cm 1cm 1.5cm}, clip, width = 0.22\textwidth]{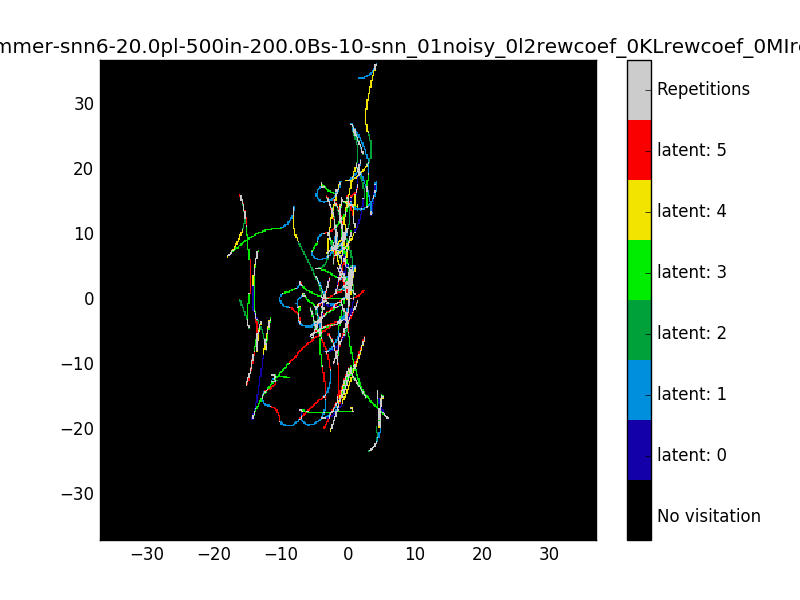}
	}
	\subfigure[Hierarchy with Bil-SNN $\alpha_H=0.01$]{
		\centering
		\label{fig:visit-hier-snnHB-1M}
		\includegraphics[trim={2cm 0.7cm 1cm 1.5cm}, clip, width = 0.22\textwidth]{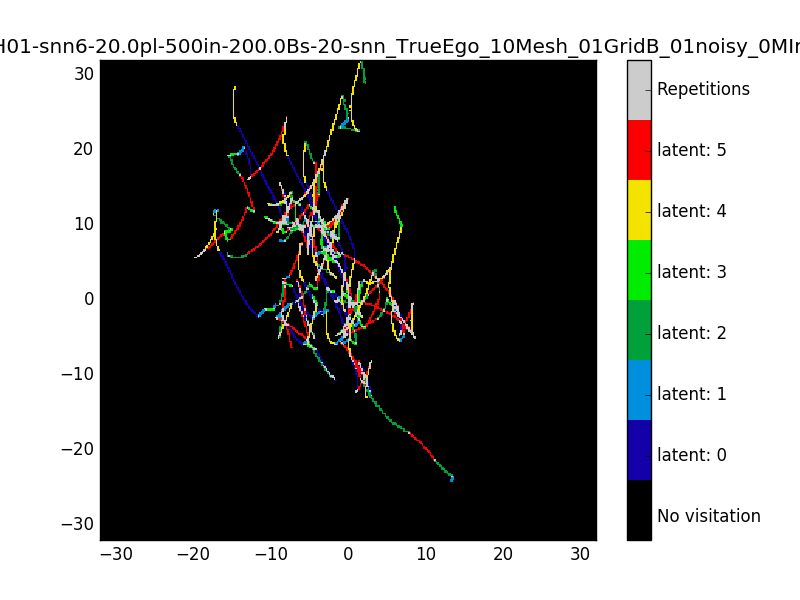}
	}
		\vspace{-8pt}
	\caption{Visitation plots for different randomly initialized architectures (one rollout of 1M steps). All axis are in scale [-30,30] and we added a zoom for the Gaussian noise to scale [-2,2]}
	\label{fig:hierarchized-exploration}
\end{figure}

On one hand, we show in Fig.\ \ref{fig:visit-trpo-1M} the exploration obtained with actions drawn from a Gaussian with $\mu=0$ and $\Sigma=I$, similar to what would happen in the first iteration of training a Multi-Layer Perceptron with normalized random initialization of the weights. This noise is relatively large as the swimmer robot has actions clipped to $[-1,1]$. Still, it does not yield good exploration as all point reached by the robot during the one million steps rollout lay within the $[-2,2]\times[-2,2]$ box around the initial position. This exploration is not enough to reach the first reward in most of the downstream sparse reward environments and will never learn, as already reported by \citet{duan2016benchmarking}. 

On the other hand, using our hierarchical structures with pretrained policies introduced in Sec.\ \ref{section:method:highlevel} yields a considerable increase in exploration, as reported in Fig.\ \ref{fig:visit-hier-multi-1M}-\ref{fig:visit-hier-snnHB-1M}. These plots also show a possible one millions steps rollout, but under a randomly initialized Manager Network, hence outputting uniformly distributed one-hot vectors every $\mathcal{T}=500$ steps. The color of every point in the CoM trajectory corresponds to the latent code that was fixed at that time-step, clearly showing the ``skill changes". In the following we describe what specific architecture was used for each plot.

The rollout in Fig.\ \ref{fig:visit-hier-multi-1M} is generated following a policy with the \textit{Multi-policy} architecture. This hierarchical architecture is our strong baseline given that it has used six times more samples for pretrain because it trained six policies independently. As explained in Sec.\ \ref{section:method:highlevel}, every $\mathcal{T}=500$ steps it uses the sampled one-hot output of the Manager Network to select one of the six pre-trained policies. We observe that the exploration given by the \textit{Multi-policy} hierarchy heavily concentrates around upward and downward motion, as is expected from the six individual pre-trained policies composing it (refer back to Fig.\ \ref{fig:visit-trpo-individuals}). 

Finally, the rollouts in Fig.\ref{fig:visit-hier-snnBil-1M} and \ref{fig:visit-hier-snnHB-1M} use our hierarchical architecture with a SNN with bilinear integration and $\alpha_H= 0$ and 0.01 respectively. As described in Sec.\ \ref{section:method:highlevel}, the placeholder that during the SNN training received the categorical latent code now receives the one-hot vector output of the Manager Network instead. The exploration obtained with SNNs yields a wider coverage of the space as the underlying policy usually has a larger span of skills. Note how most of the changes of latent code yield a corresponding change in direction. Our simple pre-train setup and the interpretability of the obtained skills are key advantages with respect to previous hierarchical structures.

\subsection{Mazes and Gather tasks}
\begin{figure}[h!]
	\centering
	\subfigure[Maze 0]{
		\centering
		\label{fig:learn-maze0}
		\includegraphics[width = 0.4\textwidth]{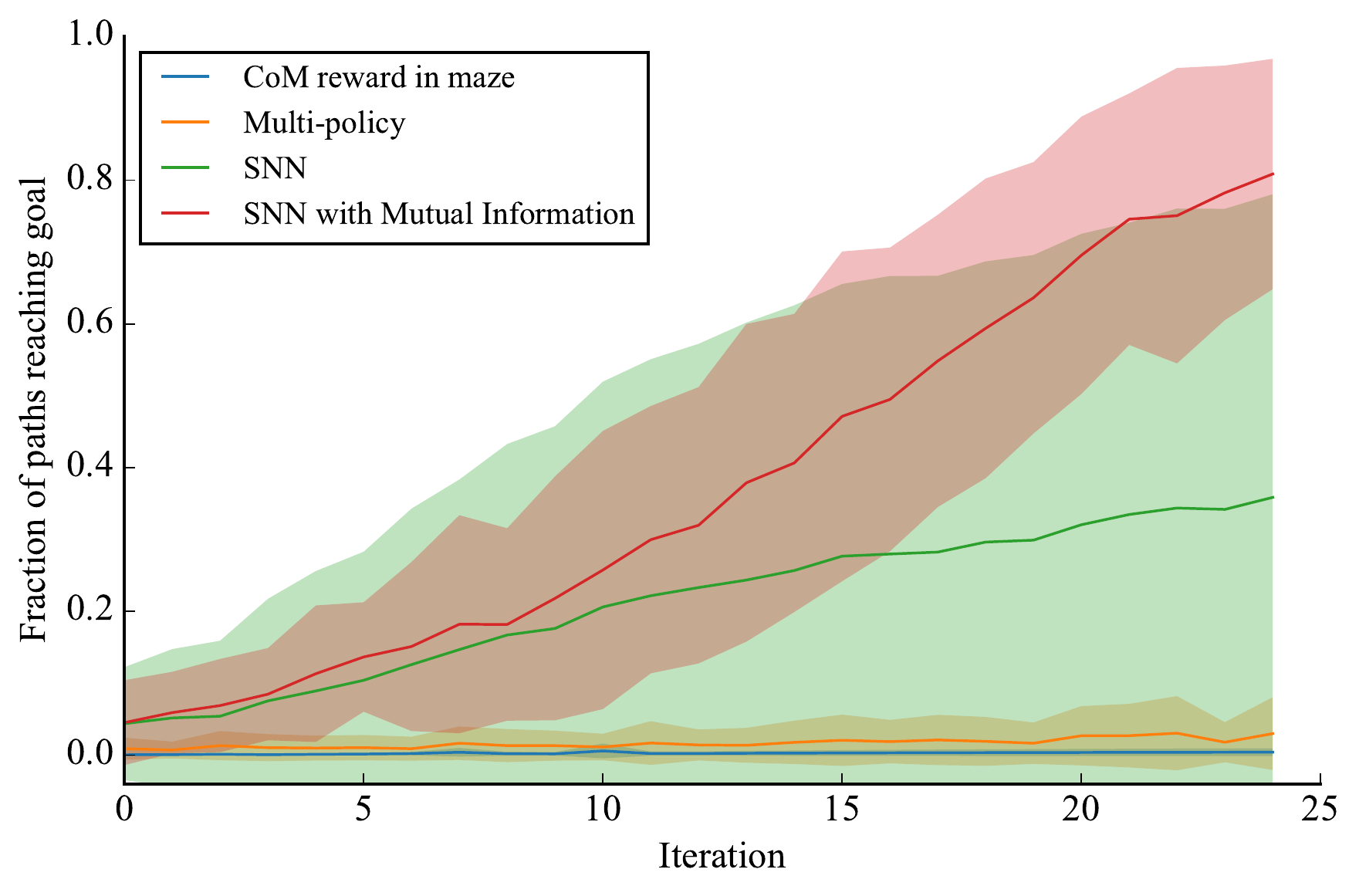}
	}
	\subfigure[Maze 1]{
		\centering
		\label{fig:learn-maze1}
		\includegraphics[width = 0.4\textwidth]{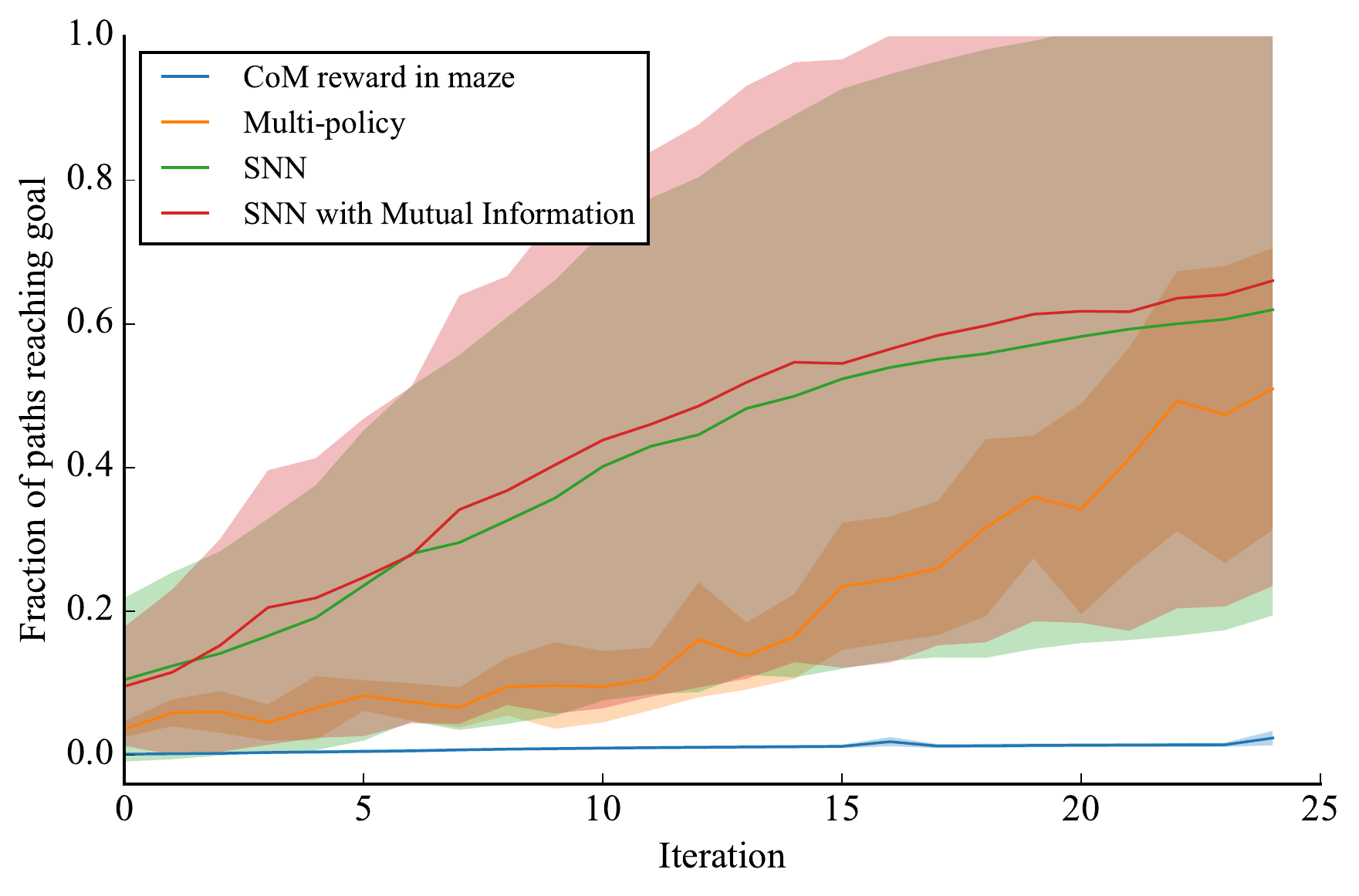}
	}
	\subfigure[Aggregated results for Mazes 2 and 3]{
		\centering
		\label{fig:learn-maze4}
		\includegraphics[width = 0.4\textwidth]{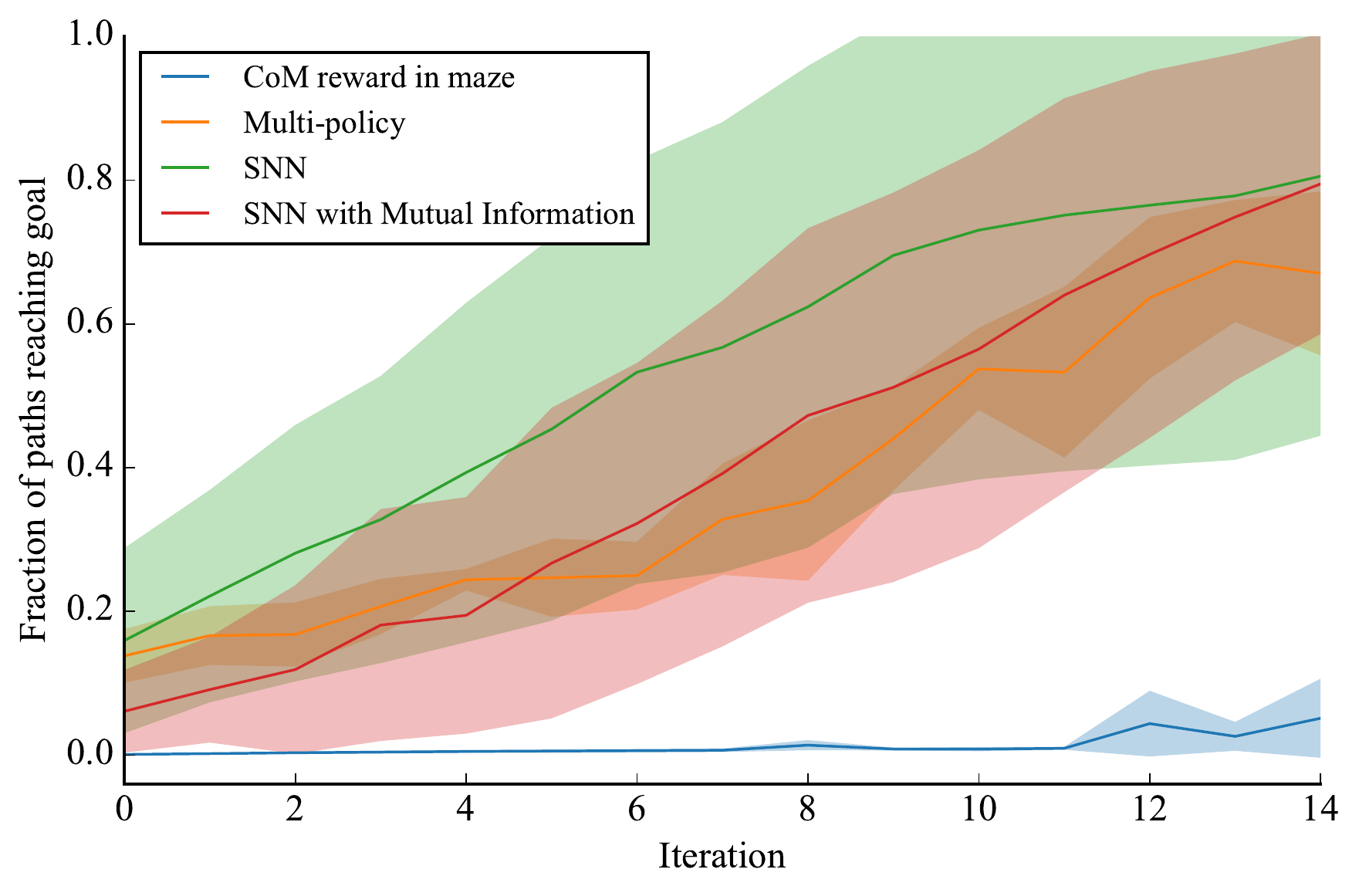}
	}
	\subfigure[Gather task]{
		\centering
		\label{fig:learn-gather}
		\includegraphics[width = 0.4\textwidth]{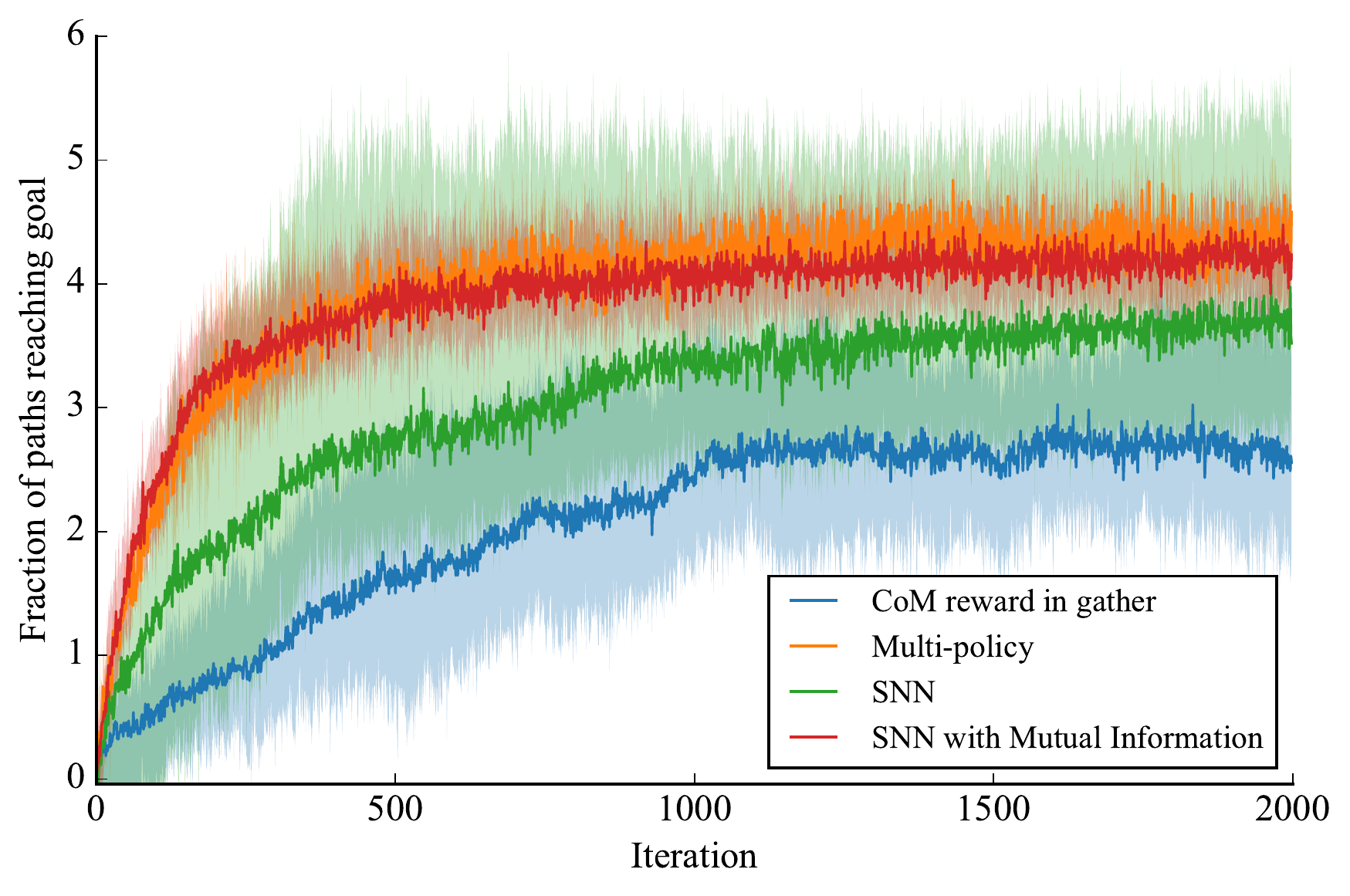}
	}
	\caption{Faster learning of the hierarchical architectures in the sparse downstream MDPs}
	\label{fig:learning-curves}
\end{figure}

In Fig.\ \ref{fig:learning-curves} we evaluate the learning curves of the different hierarchical architectures proposed. Due to the sparsity of these tasks, none can be properly solved by standard reinforcement algorithms \citep{duan2016benchmarking}. Therefore, we compare our methods against a better baseline: adding to the downstream task the same Center of Mass (CoM) proxy reward that was granted to the robot in the pre-training task. This baseline performs quite poorly in all the mazes, Fig.\ \ref{fig:learn-maze0}-\ref{fig:learn-maze4}. This is due to the long time-horizon needed to reach the goal and the associated credit assignment problem. Furthermore, the proxy reward alone does not encourage diversity of actions as the MI bonus for SNN does. 
The proposed hierarchical architectures are able to learn much faster in every new MDP as they effectively shrink the time-horizon by aggregating time-steps into useful primitives. The benefit of using SNNs with bilinear integration in the hierarchy is also clear over most mazes, although pretraining with MI bonus does not always boost performance. Observing the learned trajectories that solve the mazes, we realize that the turning or sideway motions (more present in SNNs pretrained with MI bonus) help on maze 0, but are not critical in these tasks because the robot may use a specific forward motion against the wall of the maze to reorient itself. We think that the extra skills will shine more in other tasks requiring more precise navigation. And indeed this is the case in the Gather task as seen in Fig.\ \ref{fig:learn-gather}, where not only the average return increases, but also the variance of the learning curve is lower for the algorithm using SNN pretrained with MI bonus, denoting a more consistent learning across different SNNs. For more details on the Gather task, refer to Appendix \ref{sec:gather_compare}.

It is important to mention that each curve of SNN or Multi-policy performance corresponds to a total of 10 runs: 2 random seeds for 5 different SNNs obtained with 5 random seeds in the pretrain task. There is no cherry picking of the pre-trained policy to use in the sparse environment, as done in most previous work. This also explains the high variance of the proposed hierarchical methods in some tasks. This is particularly hard on the mazes as some pretrained SNN may not have the particular motion that allows to reach the goal, so they will have a 0 reward. See Appendix \ref{sec:seed_impact} for more details.

\section{Discussion and future work}
We propose a framework for first learning a diverse set of skills using Stochastic Neural Networks trained with minimum supervision, and then utilizing these skills in a hierarchical architecture to solve challenging tasks with sparse rewards. Our framework successfully combines two parts, firstly an unsupervised procedure to learn a large span of skills using proxy rewards and secondly a hierarchical structure that encapsulates the latter span of skills and allows to re-use them in future tasks. The span of skills learning can be greatly improved by using Stochastic Neural Networks as policies and their additional expressiveness and multimodality. The bilinear integration and the mutual information bonus are key to consistently yield a wide, interpretable span of skills. As for the hierarchical structure, our experiments demonstrate it can significantly boost the exploration of an agent in a new environment and we demonstrate its relevance for solving complex tasks as mazes or gathering. 

One limitation of our current approach is the switching between skills for unstable agents, as reported in the Appendix \ref{sec:ant-failure} for the “Ant” robot. There are multiple future directions to make our framework more robust to such challenging robots, like learning a transition policy or integrating switching in the pretrain task. 
Other limitations of our current approach are having fixed sub-policies and a fixed switch time $\mathcal{T}$ during the training of downstream tasks. The first issue can be alleviated by introducing end-to-end training, for example using the new straight-through gradient estimators for Stochastic Computations Graphs with discrete latent variables \citep{jang2016gumbel, maddison2016concrete}. The second issue is not critical for static tasks like the ones used here, as studied in Appendix \ref{sec:switch_time}. But in case of becoming a bottleneck for more complex dynamic tasks, a termination policy could be learned by the Manager, similar to the option framework.
Finally, we only used feedforward architectures and hence the decision of what skill to use next only depends on the observation at the moment of switching, not using any sensory information gathered while the previous skill was active. This limitation could be eliminated by introducing a recurrent architecture at the Manager level.

\subsubsection*{Acknowledgments}
This work was supported in part by DARPA, by the Berkeley Vision and Learning Center (BVLC), by the Berkeley Artificial Intelligence Research (BAIR) laboratory, by Berkeley Deep Drive (BDD), and by ONR through a PECASE award. Carlos Florensa was also supported by a La Caixa Ph.D. Fellowship, and Yan Duan by a BAIR Fellowship and a Huawei Fellowship.

\bibliography{ref}

\begin{thebibliography}{51}
\providecommand{\natexlab}[1]{#1}
\providecommand{\url}[1]{\texttt{#1}}
\expandafter\ifx\csname urlstyle\endcsname\relax
  \providecommand{\doi}[1]{doi: #1}\else
  \providecommand{\doi}{doi: \begingroup \urlstyle{rm}\Url}\fi

\bibitem[Bacon \& Precup(2016)Bacon and Precup]{bacon2016option}
Pierre-Luc Bacon and Doina Precup.
\newblock The option-critic architecture.
\newblock \emph{arXiv preprint arXiv:1609.05140v2}, 2016.

\bibitem[Bellemare et~al.(2016)Bellemare, Srinivasan, Ostrovski, Schaul,
  Saxton, and Munos]{bellemare2016unifying}
Marc~G Bellemare, Sriram Srinivasan, Georg Ostrovski, Tom Schaul, David Saxton,
  and Remi Munos.
\newblock Unifying count-based exploration and intrinsic motivation.
\newblock \emph{Advances in Neural Information Processing Systems}, 2016.

\bibitem[Chen et~al.(2016)Chen, Duan, Houthooft, Schulman, Sutskever, and
  Abbeel]{chen2016infogan}
Xi~Chen, Yan Duan, Rein Houthooft, John Schulman, Ilya Sutskever, and Pieter
  Abbeel.
\newblock Infogan: Interpretable representation learning by information
  maximizing generative adversarial nets.
\newblock \emph{Advances in Neural Information Processing Systems}, 2016.

\bibitem[Chentanez et~al.(2004)Chentanez, Barto, and
  Singh]{chentanez2004intrinsically}
Nuttapong Chentanez, Andrew~G Barto, and Satinder~P Singh.
\newblock Intrinsically motivated reinforcement learning.
\newblock In \emph{Advances in Neural Information Processing Systems}, pp.\
  1281--1288, 2004.

\bibitem[Cho et~al.(2013)Cho, Raiko, and Ilin]{cho2013gaussian}
Kyung~Hyun Cho, Tapani Raiko, and Alexander Ilin.
\newblock Gaussian-bernoulli deep boltzmann machine.
\newblock In \emph{International Joint Conference on Neural Networks (IJCNN)},
  pp.\  1--7. IEEE, 2013.

\bibitem[Daniel et~al.(2012)Daniel, Neumann, and
  Peters]{daniel2012hierarchical}
Christian Daniel, Gerhard Neumann, and Jan Peters.
\newblock Hierarchical relative entropy policy search.
\newblock In \emph{AISTATS}, pp.\  273--281, 2012.

\bibitem[Daniel et~al.(2013)Daniel, Neumann, and Peters]{daniel2013autonomous}
Christian Daniel, Gerhard Neumann, and Jan Peters.
\newblock Autonomous reinforcement learning with hierarchical reps.
\newblock In \emph{International Joint Conference on Neural Networks}, pp.\
  1--8. IEEE, 2013.

\bibitem[Devin et~al.(2017)Devin, Gupta, Darrell, Abbeel, and
  Levine]{devin2016learning}
Coline Devin, Abhishek Gupta, Trevor Darrell, Pieter Abbeel, and Sergey Levine.
\newblock Learning modular neural network policies for multi-task and
  multi-robot transfer.
\newblock In \emph{International Conference on Robotics and Automation}, 2017.

\bibitem[Dietterich(2000)]{dietterich2000hierarchical}
Thomas~G Dietterich.
\newblock Hierarchical reinforcement learning with the maxq value function
  decomposition.
\newblock \emph{Journal of Artificial Intelligence Research}, 13:\penalty0
  227--303, 2000.

\bibitem[Duan et~al.(2016)Duan, Chen, Houthooft, Schulman, and
  Abbeel]{duan2016benchmarking}
Yan Duan, Xi~Chen, Rein Houthooft, John Schulman, and Pieter Abbeel.
\newblock Benchmarking deep reinforcement learning for continuous control.
\newblock \emph{International Conference on Machine Learning}, 2016.

\bibitem[Fukui et~al.(2016)Fukui, Park, Yang, Rohrbach, Darrell, and
  Rohrbach]{fukui2016multimodal}
Akira Fukui, Dong~Huk Park, Daylen Yang, Anna Rohrbach, Trevor Darrell, and
  Marcus Rohrbach.
\newblock Multimodal compact bilinear pooling for visual question answering and
  visual grounding.
\newblock In \emph{EMNLP}, 2016.

\bibitem[Gregor et~al.(2015)Gregor, Danihelka, Graves, Rezende, and
  Wierstra]{gregor2015draw}
Karol Gregor, Ivo Danihelka, Alex Graves, Danilo~Jimenez Rezende, and Daan
  Wierstra.
\newblock Draw: A recurrent neural network for image generation.
\newblock \emph{arXiv preprint arXiv:1502.04623}, 2015.

\bibitem[Guo et~al.(2014)Guo, Singh, Lee, Lewis, and Wang]{guo2014deep}
Xiaoxiao Guo, Satinder Singh, Honglak Lee, Richard~L Lewis, and Xiaoshi Wang.
\newblock Deep learning for real-time atari game play using offline monte-carlo
  tree search planning.
\newblock In \emph{Advances in Neural Information Processing Systems}, pp.\
  3338--3346, 2014.

\bibitem[Heess et~al.(2015)Heess, Wayne, Silver, Lillicrap, Erez, and
  Tassa]{heess2015learning}
Nicolas Heess, Gregory Wayne, David Silver, Tim Lillicrap, Tom Erez, and Yuval
  Tassa.
\newblock Learning continuous control policies by stochastic value gradients.
\newblock In \emph{Advances in Neural Information Processing Systems}, pp.\
  2944--2952, 2015.

\bibitem[Heess et~al.(2016)Heess, Wayne, Tassa, Lillicrap, Riedmiller, and
  Silver]{heess2016learning}
Nicolas Heess, Greg Wayne, Yuval Tassa, Timothy Lillicrap, Martin Riedmiller,
  and David Silver.
\newblock Learning and transfer of modulated locomotor controllers.
\newblock \emph{arXiv preprint arXiv:1610.05182}, 2016.

\bibitem[Hinton(2002)]{hinton2002training}
Geoffrey~E Hinton.
\newblock Training products of experts by minimizing contrastive divergence.
\newblock \emph{Neural Computation}, 14\penalty0 (8):\penalty0 1771--1800,
  2002.

\bibitem[Hinton et~al.(2006)Hinton, Osindero, and Teh]{hinton2006fast}
Geoffrey~E Hinton, Simon Osindero, and Yee-Whye Teh.
\newblock A fast learning algorithm for deep belief nets.
\newblock \emph{Neural Computation}, 18\penalty0 (7):\penalty0 1527--1554,
  2006.

\bibitem[Houthooft et~al.(2016)Houthooft, Chen, Duan, Schulman, De~Turck, and
  Abbeel]{houthooft2016variational}
Rein Houthooft, Xi~Chen, Yan Duan, John Schulman, Filip De~Turck, and Pieter
  Abbeel.
\newblock Variational information maximizing exploration.
\newblock \emph{Advances in Neural Information Processing Systems}, 2016.

\bibitem[Jang et~al.(2017)Jang, Gu, and Poole]{jang2016gumbel}
Eric Jang, Shixiang Gu, and Ben Poole.
\newblock Categorical reparameterization with gumbel-softmax.
\newblock In \emph{International Conference on Learning Representations}, 2017.

\bibitem[Konidaris \& Barto(2007)Konidaris and Barto]{konidaris2007building}
George Konidaris and Andrew~G Barto.
\newblock Building portable options: Skill transfer in reinforcement learning.
\newblock In \emph{IJCAI}, volume~7, pp.\  895--900, 2007.

\bibitem[Konidaris et~al.(2011)Konidaris, Kuindersma, Grupen, and
  Barto]{konidaris2011autonomous}
George Konidaris, Scott Kuindersma, Roderic~A Grupen, and Andrew~G Barto.
\newblock Autonomous skill acquisition on a mobile manipulator.
\newblock In \emph{AAAI}, 2011.

\bibitem[Lazaric \& Ghavamzadeh(2010)Lazaric and
  Ghavamzadeh]{lazaric2010bayesian}
Alessandro Lazaric and Mohammad Ghavamzadeh.
\newblock Bayesian multi-task reinforcement learning.
\newblock In \emph{27th International Conference on Machine Learning}, pp.\
  599--606. Omnipress, 2010.

\bibitem[Le~Roux \& Bengio(2008)Le~Roux and Bengio]{le2008representational}
Nicolas Le~Roux and Yoshua Bengio.
\newblock Representational power of restricted boltzmann machines and deep
  belief networks.
\newblock \emph{Neural Computation}, 20\penalty0 (6):\penalty0 1631--1649,
  2008.

\bibitem[Levine et~al.(2016)Levine, Finn, Darrell, and Abbeel]{levine2016end}
Sergey Levine, Chelsea Finn, Trevor Darrell, and Pieter Abbeel.
\newblock End-to-end training of deep visuomotor policies.
\newblock \emph{Journal of Machine Learning Research}, 17\penalty0
  (39):\penalty0 1--40, 2016.

\bibitem[Lillicrap et~al.(2015)Lillicrap, Hunt, Pritzel, Heess, Erez, Tassa,
  Silver, and Wierstra]{lillicrap2015continuous}
Timothy~P Lillicrap, Jonathan~J Hunt, Alexander Pritzel, Nicolas Heess, Tom
  Erez, Yuval Tassa, David Silver, and Daan Wierstra.
\newblock Continuous control with deep reinforcement learning.
\newblock \emph{arXiv preprint arXiv:1509.02971}, 2015.

\bibitem[Maddison et~al.(2017)Maddison, Mnih, and Teh]{maddison2016concrete}
Chris~J Maddison, Andriy Mnih, and Yee~Whye Teh.
\newblock The concrete distribution: A continuous relaxation of discrete random
  variables.
\newblock In \emph{International Conference on Learning Representations}, 2017.

\bibitem[Mannor et~al.(2004)Mannor, Menache, Hoze, and
  Klein]{mannor2004dynamic}
Shie Mannor, Ishai Menache, Amit Hoze, and Uri Klein.
\newblock Dynamic abstraction in reinforcement learning via clustering.
\newblock In \emph{21st International Conference on Machine Learning}, pp.\
  ~71. ACM, 2004.

\bibitem[Mnih et~al.(2015)Mnih, Kavukcuoglu, Silver, Rusu, Veness, Bellemare,
  Graves, Riedmiller, Fidjeland, Ostrovski, et~al.]{mnih2015human}
Volodymyr Mnih, Koray Kavukcuoglu, David Silver, Andrei~A Rusu, Joel Veness,
  Marc~G Bellemare, Alex Graves, Martin Riedmiller, Andreas~K Fidjeland, Georg
  Ostrovski, et~al.
\newblock Human-level control through deep reinforcement learning.
\newblock \emph{Nature}, 518\penalty0 (7540):\penalty0 529--533, 2015.

\bibitem[Mnih et~al.(2016)Mnih, Agapiou, Osindero, Graves, Vinyals,
  Kavukcuoglu, et~al.]{mnih2016strategic}
Volodymyr Mnih, John Agapiou, Simon Osindero, Alex Graves, Oriol Vinyals, Koray
  Kavukcuoglu, et~al.
\newblock Strategic attentive writer for learning macro-actions.
\newblock In \emph{Advances in Neural Information Processing Systems}, 2016.

\bibitem[Neal(1990)]{neal1990learning}
Radford~M Neal.
\newblock Learning stochastic feedforward networks.
\newblock \emph{Department of Computer Science, University of Toronto}, 1990.

\bibitem[Neal(1992)]{neal1992connectionist}
Radford~M Neal.
\newblock Connectionist learning of belief networks.
\newblock \emph{Artificial intelligence}, 56\penalty0 (1):\penalty0 71--113,
  1992.

\bibitem[Osband et~al.(2014)Osband, Van~Roy, and Wen]{osband2014generalization}
Ian Osband, Benjamin Van~Roy, and Zheng Wen.
\newblock Generalization and exploration via randomized value functions.
\newblock \emph{arXiv preprint arXiv:1402.0635}, 2014.

\bibitem[Parr \& Russell(1998)Parr and Russell]{parr1998reinforcement}
Ronald Parr and Stuart Russell.
\newblock Reinforcement learning with hierarchies of machines.
\newblock \emph{Advances in Neural Information Processing Systems}, pp.\
  1043--1049, 1998.

\bibitem[Ranchod et~al.(2015)Ranchod, Rosman, and
  Konidaris]{ranchod2015nonparametric}
Pravesh Ranchod, Benjamin Rosman, and George Konidaris.
\newblock Nonparametric bayesian reward segmentation for skill discovery using
  inverse reinforcement learning.
\newblock In \emph{International Conference on Intelligent Robots and Systems},
  pp.\  471--477. IEEE, 2015.

\bibitem[Schaal et~al.(2005)Schaal, Peters, Nakanishi, and
  Ijspeert]{schaal2005learning}
Stefan Schaal, Jan Peters, Jun Nakanishi, and Auke Ijspeert.
\newblock Learning movement primitives.
\newblock In \emph{Robotics Research. The Eleventh International Symposium},
  pp.\  561--572, 2005.

\bibitem[Schmidhuber(1991)]{schmidhuber1991curious}
J{\"u}rgen Schmidhuber.
\newblock Curious model-building control systems.
\newblock In \emph{International Joint Conference on Neural Networks}, pp.\
  1458--1463. IEEE, 1991.

\bibitem[Schmidhuber(2010)]{schmidhuber2010formal}
J{\"u}rgen Schmidhuber.
\newblock Formal theory of creativity, fun, and intrinsic motivation
  (1990--2010).
\newblock \emph{IEEE Transactions on Autonomous Mental Development}, 2\penalty0
  (3):\penalty0 230--247, 2010.

\bibitem[Schulman et~al.(2015)Schulman, Levine, Abbeel, Jordan, and
  Moritz]{Schulman15TRPO}
J.~Schulman, S.~Levine, P.~Abbeel, M.~I. Jordan, and P.~Moritz.
\newblock Trust region policy optimization.
\newblock In \emph{International Conference on Learning Representations}, pp.\
  1889--1897, 2015.

\bibitem[Schulman et~al.(2016)Schulman, Moritz, Levine, Jordan, and
  Abbeel]{schulman2016high}
John Schulman, Philipp Moritz, Sergey Levine, Michael Jordan, and Pieter
  Abbeel.
\newblock High-dimensional continuous control using generalized advantage
  estimation.
\newblock In \emph{International Conference on Learning Representations}, 2016.

\bibitem[Silver et~al.(2016)Silver, Huang, Maddison, Guez, Sifre, Van
  Den~Driessche, Schrittwieser, Antonoglou, Panneershelvam, Lanctot,
  et~al.]{silver2016mastering}
David Silver, Aja Huang, Chris~J Maddison, Arthur Guez, Laurent Sifre, George
  Van Den~Driessche, Julian Schrittwieser, Ioannis Antonoglou, Veda
  Panneershelvam, Marc Lanctot, et~al.
\newblock Mastering the game of go with deep neural networks and tree search.
\newblock \emph{Nature}, 529\penalty0 (7587):\penalty0 484--489, 2016.

\bibitem[{\c{S}}im{\c{s}}ek et~al.(2005){\c{S}}im{\c{s}}ek, Wolfe, and
  Barto]{csimcsek2005identifying}
{\"O}zg{\"u}r {\c{S}}im{\c{s}}ek, Alicia~P Wolfe, and Andrew~G Barto.
\newblock Identifying useful subgoals in reinforcement learning by local graph
  partitioning.
\newblock In \emph{Proceedings of the 22nd international conference on Machine
  learning}, pp.\  816--823. ACM, 2005.

\bibitem[Smolensky(1986)]{smolensky1986information}
Paul Smolensky.
\newblock Information processing in dynamical systems: Foundations of harmony
  theory.
\newblock Technical report, DTIC Document, 1986.

\bibitem[Stolle \& Precup(2002)Stolle and Precup]{stolle2002learning}
Martin Stolle and Doina Precup.
\newblock Learning options in reinforcement learning.
\newblock In \emph{International Symposium on Abstraction, Reformulation, and
  Approximation}, pp.\  212--223, 2002.

\bibitem[Sutton et~al.(1999)Sutton, Precup, and Singh]{sutton1999between}
Richard~S Sutton, Doina Precup, and Satinder Singh.
\newblock Between mdps and semi-mdps: A framework for temporal abstraction in
  reinforcement learning.
\newblock \emph{Artificial intelligence}, 112\penalty0 (1):\penalty0 181--211,
  1999.

\bibitem[Tang \& Salakhutdinov(2013)Tang and Salakhutdinov]{Tang2014_FSNN}
Yichuan Tang and Ruslan~R Salakhutdinov.
\newblock Learning stochastic feedforward neural networks.
\newblock In \emph{Advances in Neural Information Processing Systems 26}, pp.\
  530--538, 2013.

\bibitem[Taylor \& Stone(2009)Taylor and Stone]{taylor2009transfer}
Matthew~E Taylor and Peter Stone.
\newblock Transfer learning for reinforcement learning domains: A survey.
\newblock \emph{Journal of Machine Learning Research}, 10:\penalty0 1633--1685,
  2009.

\bibitem[Todorov \& Li(2005)Todorov and Li]{todorov2005ilqg}
Emanuel Todorov and Weiwei Li.
\newblock A generalized iterative lqg method for locally-optimal feedback
  control of constrained nonlinear stochastic systems.
\newblock In \emph{American Control Conference, 2005. Proceedings of the 2005},
  pp.\  300--306. IEEE, 2005.

\bibitem[Vigorito \& Barto(2010)Vigorito and Barto]{vigorito2010intrinsically}
Christopher~M Vigorito and Andrew~G Barto.
\newblock Intrinsically motivated hierarchical skill learning in structured
  environments.
\newblock \emph{IEEE Transactions on Autonomous Mental Development}, 2\penalty0
  (2):\penalty0 132--143, 2010.

\bibitem[Watter et~al.(2015)Watter, Springenberg, Boedecker, and
  Riedmiller]{watter2015embed}
Manuel Watter, Jost Springenberg, Joschka Boedecker, and Martin Riedmiller.
\newblock Embed to control: A locally linear latent dynamics model for control
  from raw images.
\newblock In \emph{Advances in Neural Information Processing Systems}, pp.\
  2746--2754, 2015.

\bibitem[Wilson et~al.(2007)Wilson, Fern, Ray, and Tadepalli]{wilson2007multi}
Aaron Wilson, Alan Fern, Soumya Ray, and Prasad Tadepalli.
\newblock Multi-task reinforcement learning: a hierarchical bayesian approach.
\newblock In \emph{Proceedings of the 24th international conference on Machine
  learning}, pp.\  1015--1022. ACM, 2007.

\bibitem[Wu et~al.(2016)Wu, Zhang, Zhang, Bengio, and
  Salakhutdinov]{wu2016multiplicative}
Yuhuai Wu, Saizheng Zhang, Ying Zhang, Yoshua Bengio, and Ruslan Salakhutdinov.
\newblock On multiplicative integration with recurrent neural networks.
\newblock In \emph{Advances in Neural Information Processing Systems}, 2016.

\end{thebibliography}
\bibliographystyle{iclr2017_conference}

\appendix
\section{Hyperparameters}
\label{sec:hyper}
All policies are trained with TRPO with step size $0.01$ and discount $0.99$. All neural networks (each of the Multi-policy ones, the SNN and the Manager Network) have 2 layers of 32 hidden units. For the SNN training, the mesh density used to grid the $(x, y)$ space and give the MI bonus is 10 divisions/unit. The number of skills trained (ie dimension of latent variable in the SNN or number of independently trained policies in the Mulit-policy setup) is 6. The batch size and the maximum path length for the pre-train task are also the ones used in the benchmark \citep{duan2016benchmarking}: 50,000 and 500 respectively. For the downstream tasks, see Tab.\ \ref{tab:params}.

\begin{table}[h!]
\centering
\begin{tabular}[t]{l|c|c}
Parameter & Mazes & Food Gather \\
\hline
Batch size & 1M & 100k \\
Maximum path length & 10k & 5k \\
Switch time $\mathcal{T}$ & 500 & 10
\end{tabular}
\caption{Parameters of algorithms for downstream tasks, measured in number of time-steps of the low level control}
\label{tab:params}
\end{table}

\section{Results for Gather with Benchmark settings}
\label{sec:gather_compare}
To fairly compare our methods to previous work on the downstream Gather environment, we report here our results in the exact settings used in \citet{duan2016benchmarking}: maximum path length of 500 and batch-size of 50k. Our SNN hierarchical approach outperforms state-of-the-art intrinsic motivation results like VIME \citep{houthooft2016variational}.

The baseline of having a Center of Mass speed intrinsic reward in the task happens to be stronger than expected. This is due to two factors. First, the task offers a not-so-sparse reward (green and red balls may lie quite close to the robot), which can be easily reached by an agent intrinsically motivated to move. Second, the limit of 500 steps makes the original task much simpler in the sense that the agent only needs to learn how to reach the nearest green ball, as it won't have time to reach anything else. Therefore there is no actual need of skills to re-orient or properly navigate the environment, making the hierarchy useless. This is why when increasing the time-horizon to 5,000 steps, the hierarchy shines much more, as can also be seen in the videos.

\begin{figure}[h!]
	\centering
	\label{fig:learn-gather-benchmark}
	\includegraphics[width = 0.5\textwidth]{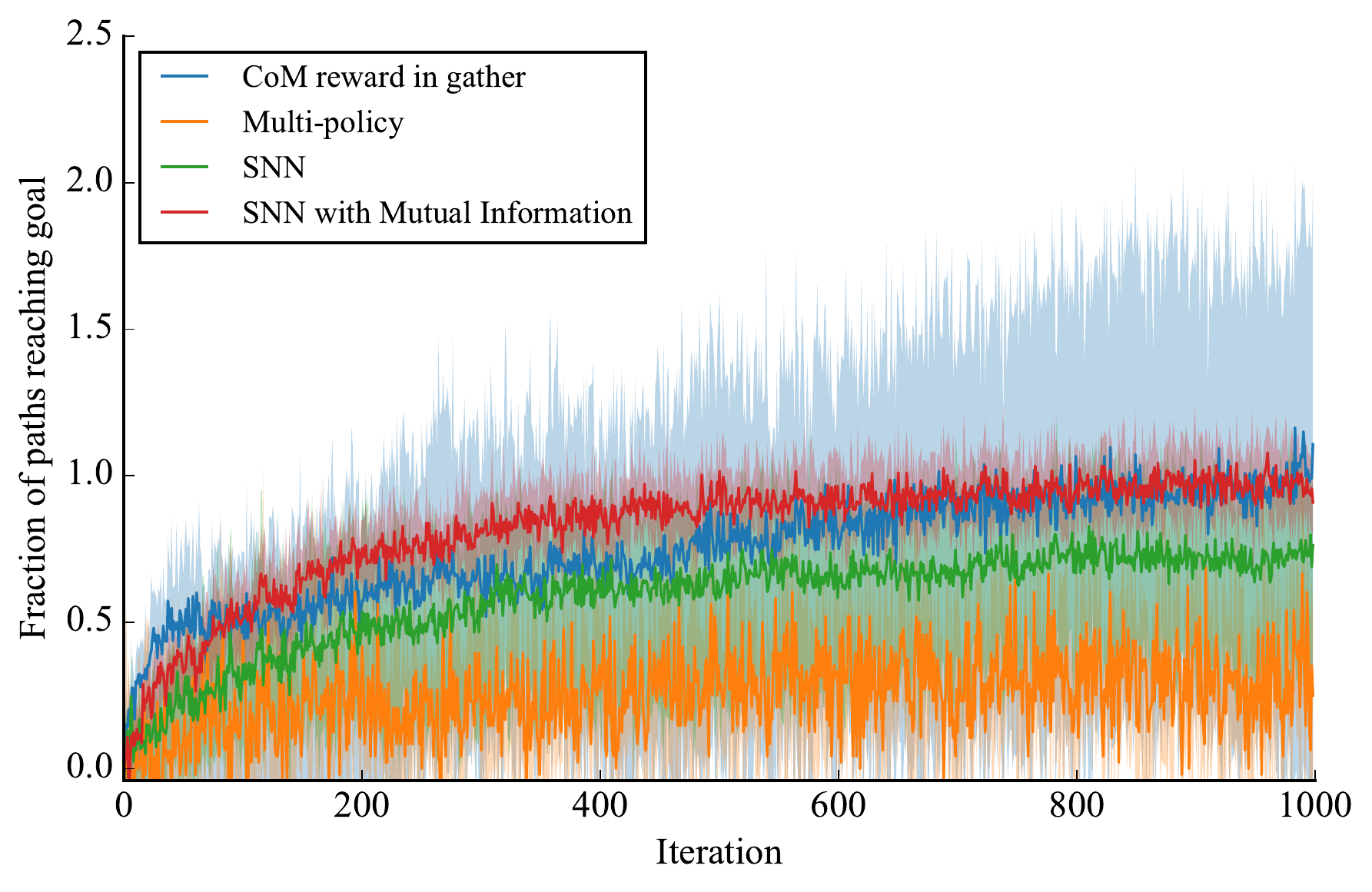}
	\caption{Results for Gather environment in the benchmark settings}
\end{figure}

\section{Snake}
\label{sec:snake}

\begin{wrapfigure}{r}{2.5cm}
\vspace{-20pt}
\includegraphics[trim={2cm 3cm 1cm 1.5cm}, clip, width = 2.5cm]{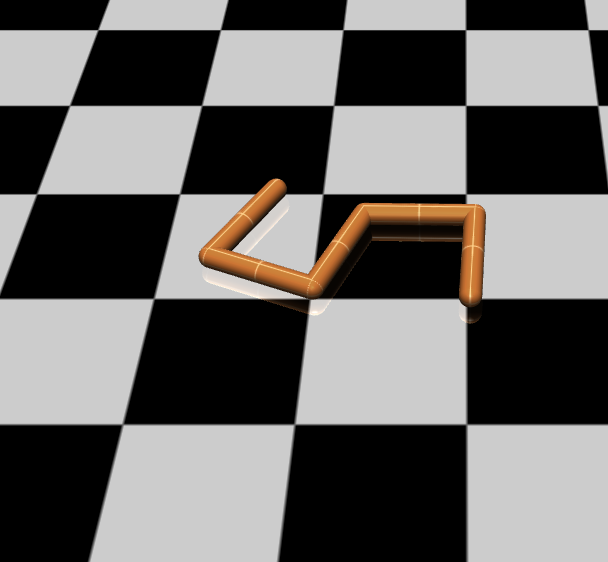}
\vspace{-15pt}
\caption{Snake}
\label{fig:snake-curl}
\end{wrapfigure}

In this section we study how our method scales to a more complex robot. We repeat most experiments from Section 7 with the "Snake" agent, a 5-link robot depicted in Fig.\ \ref{fig:snake-curl}. Compared to Swimmer, the action dimension doubles and the state-space increases by 50\%. We also perform a further analysis on the relevance of the switch time $\mathcal{T}$ and end up by reporting the performance variation among different pretrained SNNs on the hierarchical tasks.

\subsection{Skill learning in pretrain}
\label{sec:snake-pretrain}
\begin{figure}[h!]
	\centering
	\subfigure[seed 10]{
		\centering
		\label{fig:visit-snn-snake-10}
		\includegraphics[trim={2cm 0.7cm 5cm 1.5cm}, clip, width = 0.16\textwidth]{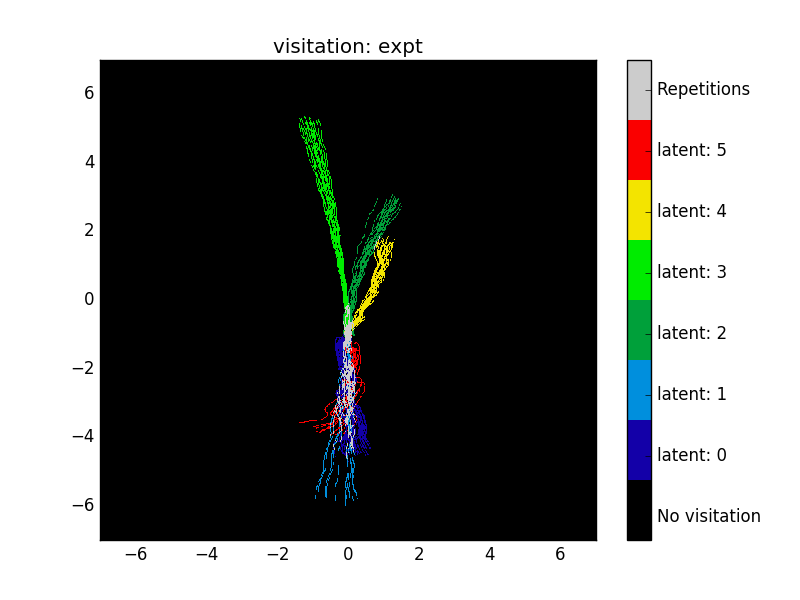}
	}
	\subfigure[seed 20]{
		\centering
		\label{fig:visit-snn-snake-20}
		\includegraphics[trim={2cm 0.7cm 5cm 1.5cm}, clip, width = 0.16\textwidth]{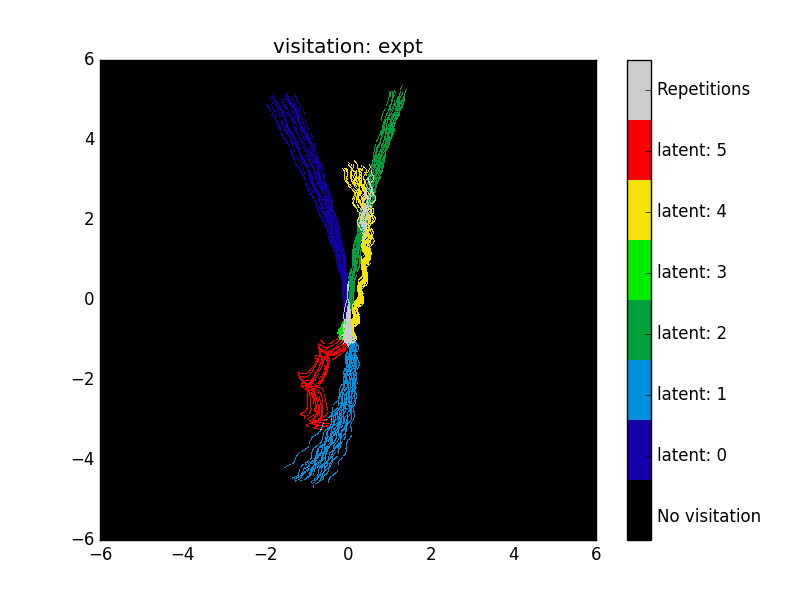}
	}
	\subfigure[seed 30]{
		\centering
		\label{fig:visit-snn-snake-30}
		\includegraphics[trim={2cm 0.7cm 5cm 1.5cm}, clip, width = 0.16\textwidth]{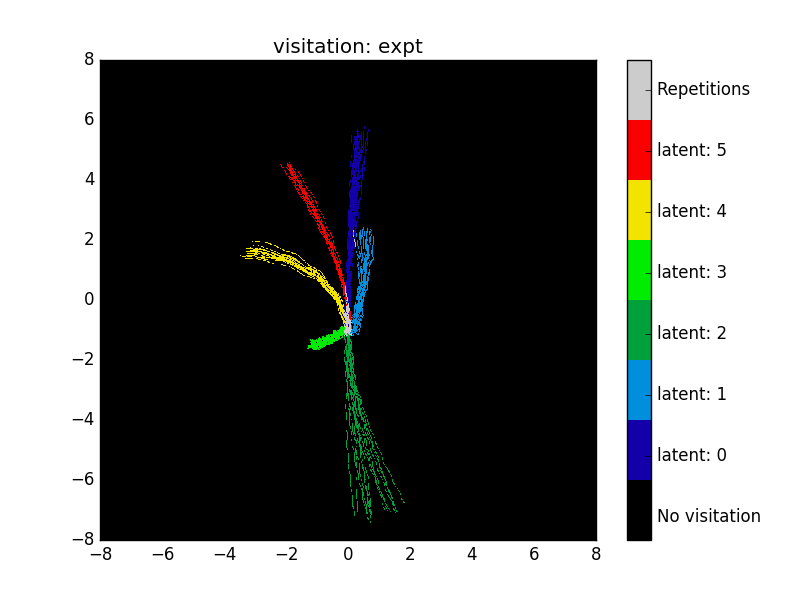}
	}
	\subfigure[seed 40]{
		\centering
		\label{fig:visit-snn-snake-40}
		\includegraphics[trim={2cm 0.7cm 5cm 1.5cm}, clip, width = 0.16\textwidth]{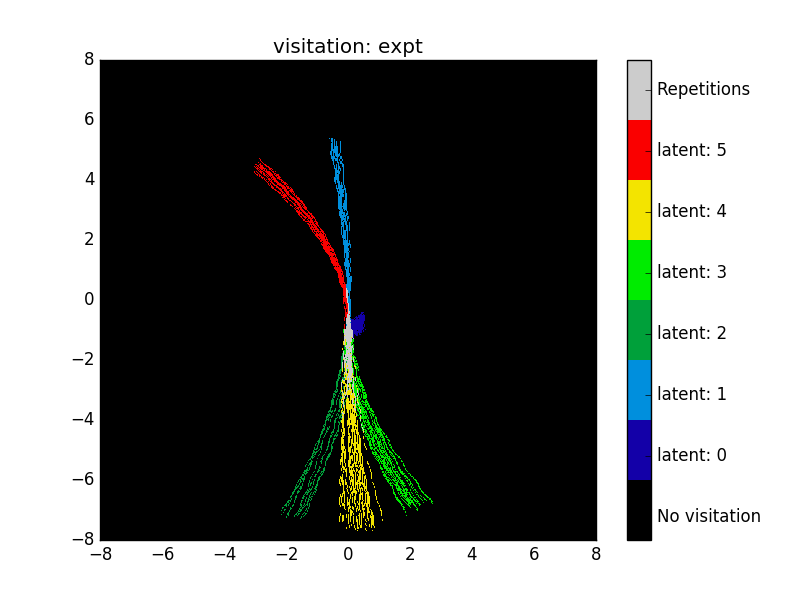}
	}
	\subfigure[seed 50]{
		\centering
		\label{fig:visit-snn-snake-50}
		\includegraphics[trim={2cm 0.7cm 5cm 1.5cm}, clip, width = 0.16\textwidth]{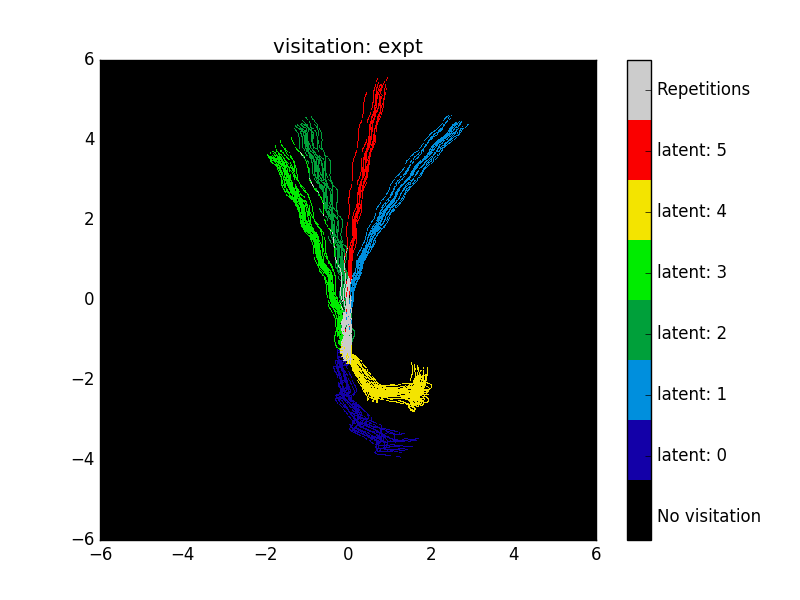}
	}
	\subfigure{
    	\centering
    	\includegraphics[trim={15.5cm 0.7cm 1cm 1.5cm}, clip, width = 0.07\textwidth, height = 0.155\textwidth]{Figures/egoSnake64-snn_005MI_5grid_6lat_categorical_bil_0050_visitation.png}
	}
	\caption{Span of skills learned using different random seeds to pretraining a SNN with MI $= 0.05$}
	\label{fig:visitation-snn-egoSnake}
\end{figure}

The Snake robot can learn a large span of skills as consistently as Swimmer. We report in Figs.\ \ref{fig:visit-snn-snake-10}-\ref{fig:visit-snn-snake-50} the visitation plots obtained for five different random seeds in our SNN pretraining algorithm from Secs.\ \ref{section:method:pretraining}-\ref{section:method:inforeg}, with Mutual Information bonus of 0.05. The impact of the different spans of skills on the later hierarchical training is discussed in Sec.\ \ref{sec:seed_impact}.

\subsection{Hierarchical use of skills in Maze and Gather}
\label{sec:snake-hierarchy}
\begin{figure}[h!]
	\centering
	\subfigure[Maze 0 with Snake]{
		\centering
		\label{fig:egoSnake-maze0}
		\includegraphics[width = 0.45\textwidth]{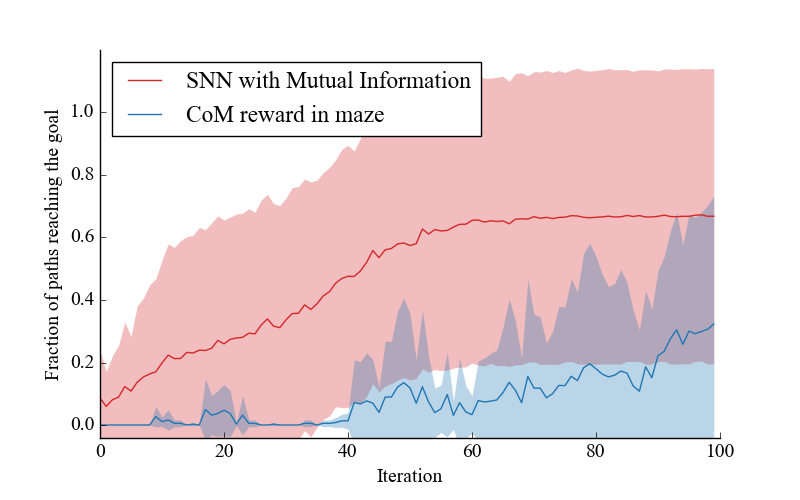}
	}
	\subfigure[Gather task with Snake]{
		\centering
		\label{fig:egoSnake-gather}
		\includegraphics[width = 0.45\textwidth]{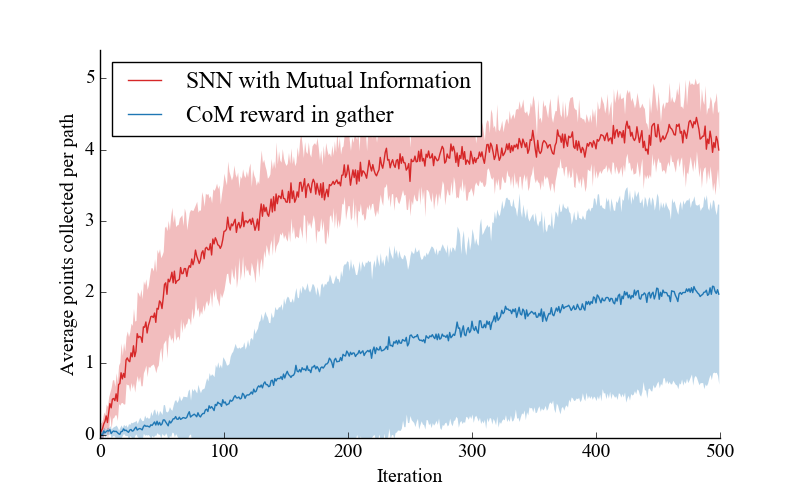}
	}
	\caption{Faster learning of our hierarchical architecture in sparse downstream MDPs with Snake}
	\label{fig:egoSnake-learning-curves}
\end{figure}

Here we evaluate the performance of our hierarchical architecture to solve with Snake the sparse tasks Maze 0 and Gather from Sec.\ \ref{sec:experiments}. Given the larger size of the robot, we have also increased the size of the Maze and the Gather task, from 2 to 7 and from 6 to 10 respectively. The maximum path length is also increased in the same proportion, but not the batch size nor the switch time $\mathcal{T}$ (see Sec.\ \ref{sec:switch_time} for an analysis on $\mathcal{T}$). Despite the tasks being harder, our approach still achieves good results, and Figs.\ \ref{fig:egoSnake-maze0}-\ref{fig:egoSnake-gather}, clearly shows how it outperforms the baseline of having the intrinsic motivation of the Center of Mass in the hierarchical task.

\subsection{Analysis of the switch time $\mathcal{T}$}
\label{sec:switch_time}
\begin{figure}[h!]
	\centering
	\subfigure[Gather task with size 10]{
		\centering
		\label{fig:egoSnake-gather-range10-switch}
		\includegraphics[trim={0cm 0cm 0cm 1.5cm}, clip, width = 0.4\textwidth]{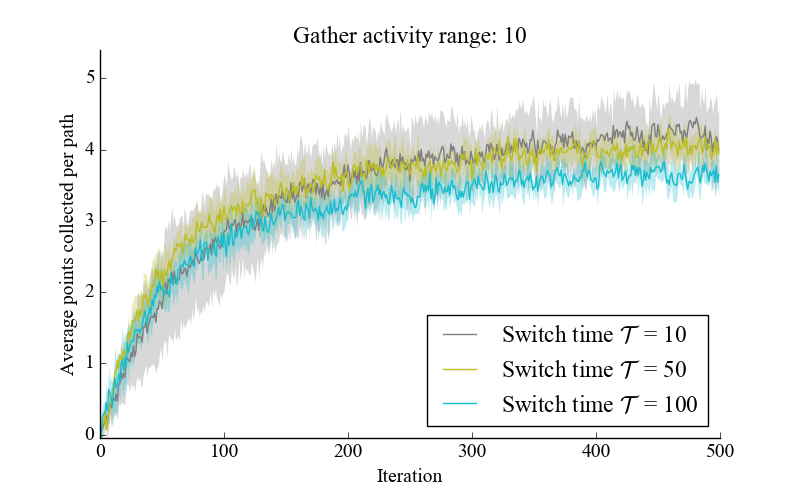}
	}
	\subfigure[Gather task with size 15]{
		\centering
		\label{fig:egoSnake-gather-range15-switch}
		\includegraphics[trim={0cm 0cm 0cm 1.5cm}, clip, width = 0.4\textwidth]{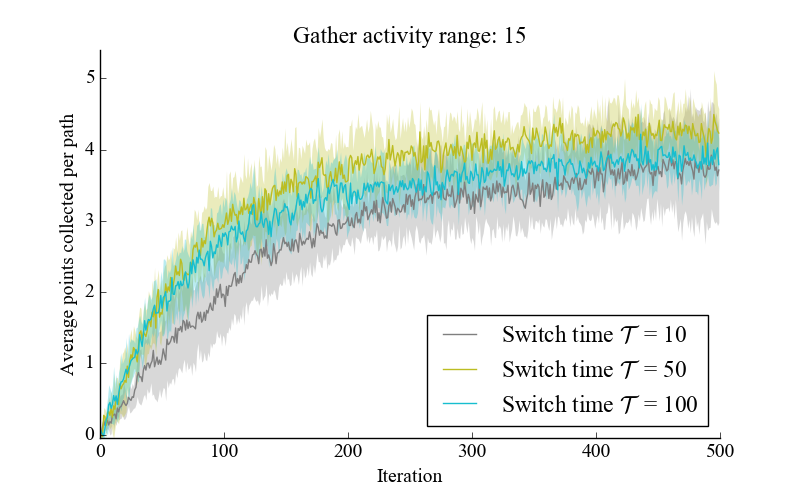}
	}
	\subfigure[Maze 0 task with size 7]{
		\centering
		\label{fig:egoSnake-maze-range7-switch}
		\includegraphics[trim={0cm 0cm 0cm 1.5cm}, clip, width = 0.4\textwidth]{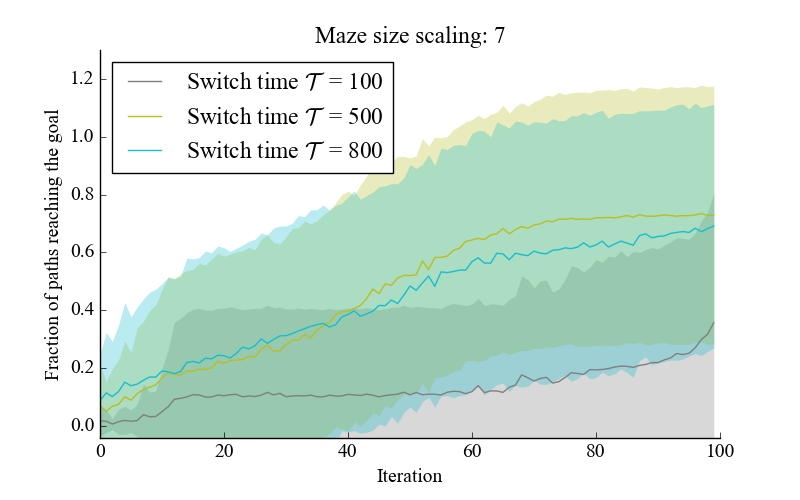}
	}
	\subfigure[Maze 0 task with size 9]{
		\centering
		\label{fig:egoSnake-maze-range9-switch}
		\includegraphics[trim={0cm 0cm 0cm 1.5cm}, clip, width = 0.4\textwidth]{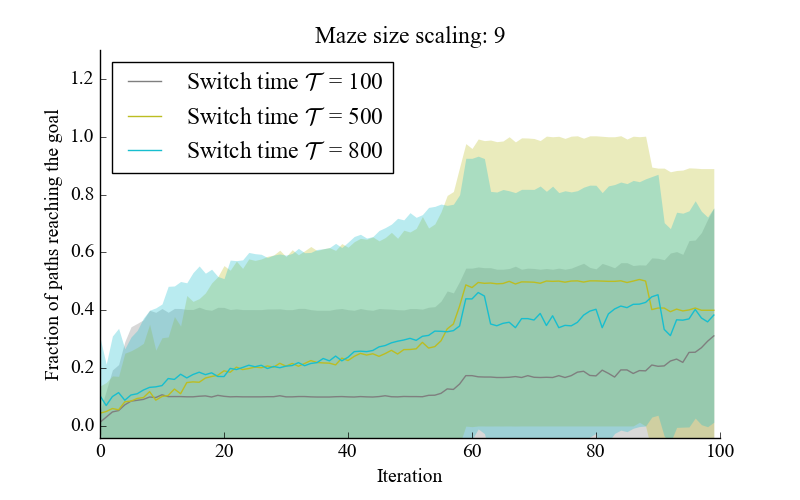}
	}
    \vspace{-10pt}    
	\caption{Mild effect of switch time $\mathcal{T}$ on different sizes of Gather and Maze 0}
	\label{fig:egoSnake-learning-curves-switch}
\end{figure}

The switch time $\mathcal{T}$ does not critically affect the performance for these static tasks, where the robot is not required to react fast and precisely to changes in the environment. The performance of very different $\mathcal{T}$ is reported for the Gather task of sizes 10 and 15 in Figs.\ \ref{fig:egoSnake-gather-range10-switch}-\ref{fig:egoSnake-gather-range15-switch}. A smaller environment implies having red/green balls closer to each other, and hence more precise navigation is expected to achieve higher scores. In our framework, lower switch time $\mathcal{T}$ allows faster changes between skills, therefore explaining the slightly better performance of $\mathcal{T}=10$ or 50 over 100 for the Gather task of size 10 (Fig.\ \ref{fig:egoSnake-gather-range10-switch}). On the other hand, larger environments mean more sparcity as the same number of balls is uniformly initialized in a wider space. In such case, longer commitment to every skill is expected to improve the exploration range. This explains the slower learning of $\mathcal{T}=10$ in the Gather task of size 15 (Fig.\ \ref{fig:egoSnake-gather-range15-switch}). It is still surprising the mild changes produced by such large variations in the switch time $\mathcal{T}$ for the Gather task.

For the Maze 0 task, Figs.\ \ref{fig:egoSnake-maze-range7-switch}-\ref{fig:egoSnake-maze-range9-switch} show that the difference between different $\mathcal{T}$ is important but not critical. In particular, radical increases of $\mathcal{T}$ have little effect on the performance in this task because the robot will simply keep bumping against a wall until it switches latent code. This behavior is observed in the videos attached with the paper\textsuperscript{\ref{foot:video}}. The large variances observed are due to the performance of the different pretrained SNN. This is further studied in Sec.\ \ref{sec:seed_impact}.

\subsection{Impact of the pre-trained SNN on the hierarchy}
\label{sec:seed_impact}
\begin{figure}[h]
	\centering
	\subfigure[Snake on Maze 0 of size 7 and switch-time $\mathcal{T}=500$]{
		\centering
		\label{fig:egoSnake-maze0-switch500-pkl}
		\includegraphics[trim={0cm 0cm 0cm 1.5cm}, clip, width = 0.45\textwidth]{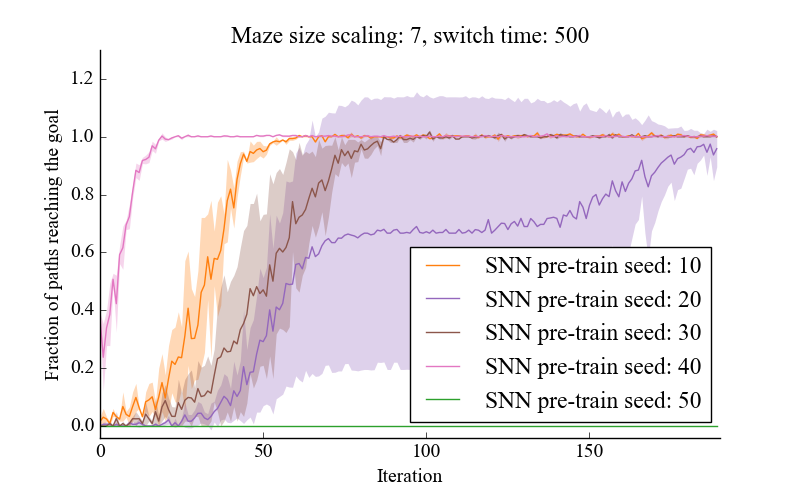}
	}
	\subfigure[Snake on Maze 0 of size 7 trained with TRPO and CoM reward]{
		\centering
		\label{fig:egoSnake-maze0-TRPO-seeds}
		\includegraphics[trim={0cm 0cm 0cm 1.5cm}, clip, width = 0.45\textwidth]{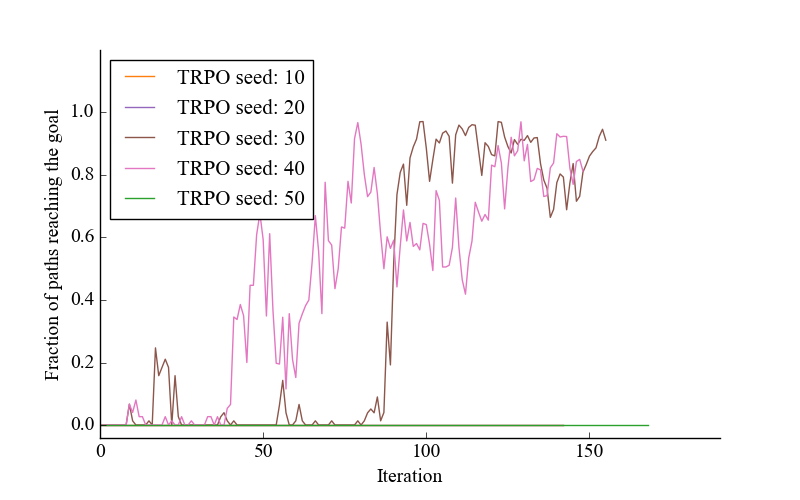}
	}	
    \vspace{-10pt}         
	\caption{Variation in performance among different pretrained SNNs and different TRPO seeds}
	\label{fig:egoSnake-learning-curves-pkl}
\end{figure}

In the previous section, the low variance of the learning curves for Gather indicates that all pretrained SNN preform equally good in this task. For Maze, the performance depends strongly on the pretrained SNN. For example we observe that the one obtained with the random seed 50 has a visitation with a weak backward span (Fig.\ \ref{fig:visit-snn-snake-50}), which happens to be critical to solve the Maze 0 (as can be seen in the videos\textsuperscript{\ref{foot:video}}). This explains the lack of learning in Fig.\ \ref{fig:egoSnake-maze0-switch500-pkl} for this particular pretrained SNN. Note that the large variability between different random seeds is even worse in the baseline of having the CoM reward in the Maze, as seen in Fig.\ \ref{fig:egoSnake-maze0-TRPO-seeds}. 3 out of 5 seeds never reach the goal and the ones that does have an unstable learning due to the long time-horizon.

\section{Ant}
\label{sec:ant}
\begin{wrapfigure}{r}{2.5cm}
\vspace{-40pt}
\includegraphics[trim={0cm 2cm 1cm 2.5cm}, clip, width = 2.5cm]{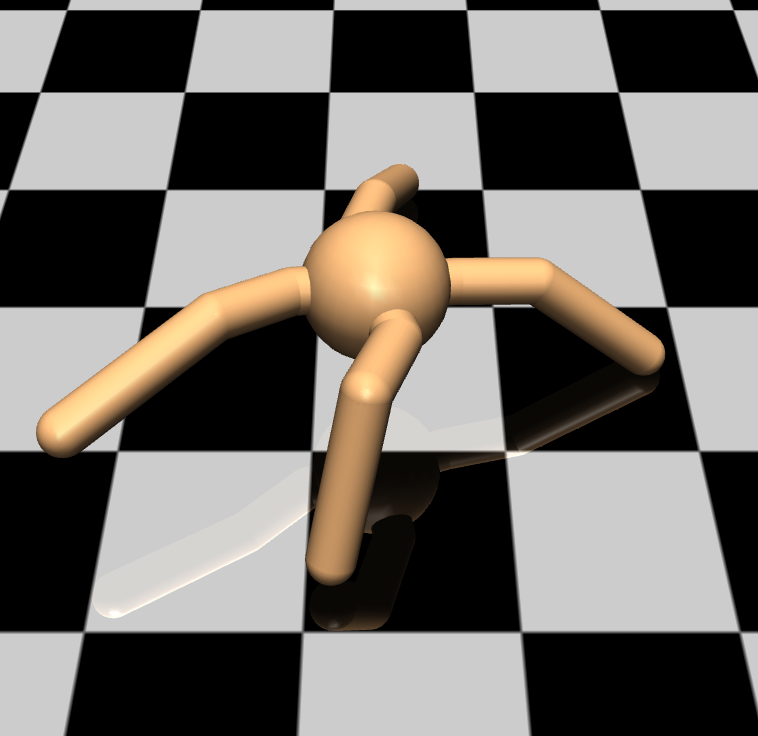}
\vspace{-20pt}
\caption{Ant}
\label{fig:ant}
\end{wrapfigure}

In this section we report the progress and failure modes of our approach with more challenging and unstable robots, like Ant (described in \cite{duan2016benchmarking}). This agent has a 27-dimensional space and an 8-dimensional action space. It is unstable in the sense that it might fall over and cannot recover. We show that despite being able to learn well differentiated skills with SNNs, switching between skills is a hard task because of the instability.

\subsection{Sill learning in pretrain}
\begin{figure}[h]
    \centering
    \includegraphics[width = 0.8\textwidth]{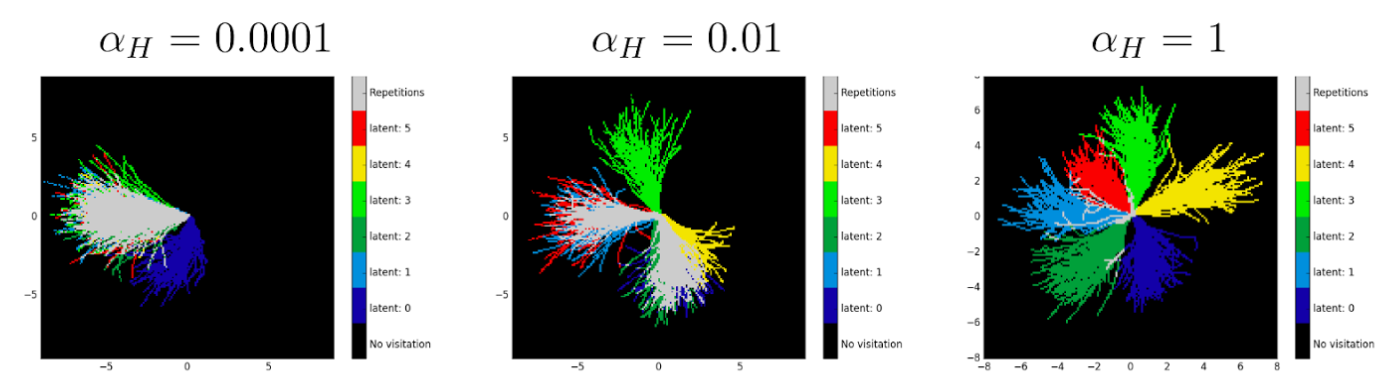}
    \caption{Pretrain visitation for Ant with different MI bonus}
    \label{fig:ant-pretrain}
\end{figure}

Our SNN pretraining step is still able to yield large span of skills covering most directions, as long as the MI bonus is well tuned, as observed in Fig.\ \ref{fig:ant-pretrain}. Videos comparing all the different gaits can be observed in the attached videos\footnote{Videos available at: \url{http://bit.ly/snn4hrl-videos}}.

\subsection{Failure modes for unstable robots}
\label{sec:ant-failure}
When switching to a new skill, the robot finds itself in a region of the state-space that might be unknown to the new skill. This increases the instability of the robot, and for example in the case of Ant this might leads to falling over and terminating the rollout. We see this in Fig.\ \ref{fig:ant-failure}, where the 5 depicted rollouts terminate because of the ant falling over. For more details on failure cases with Ant and the possible solutions, please refer to the technical report: \url{http://bit.ly/snn4hrl-antReport}

\begin{figure}[h]
    \centering
    \includegraphics[width = 0.4\textwidth]{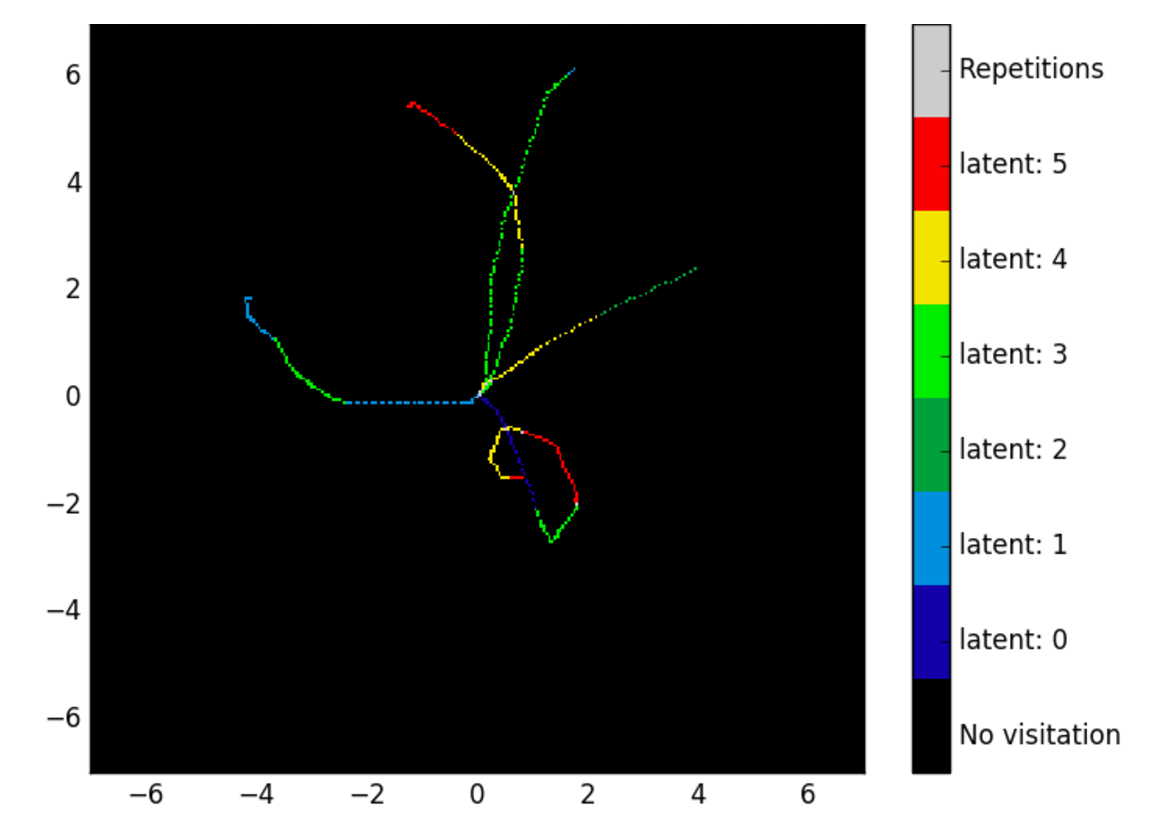}
    \caption{Failure mode of Ant: here 5 rollouts terminate in less than 6 skill switches.}
    \label{fig:ant-failure}
\end{figure}


\end{document}